%% file: root.tex
\documentclass[letterpaper,journal]{IEEEtran}
\usepackage{cite}
\usepackage{amsmath,amsfonts,amsthm}
\newtheorem{theorem}{Theorem}
\newtheorem{lemma}{Lemma}
\newtheorem{corollary}{Corollary}
\newtheorem{assumption}{Assumption}
\newtheorem{definition}{Definition}
\theoremstyle{remark}
\newtheorem{remark}{Remark}
\usepackage{algorithm}
\usepackage{array}
\newcommand{\tablefont}{\fontsize{8.5}{10}\selectfont}
\usepackage[caption=false,font=normalsize,labelfont=sf,textfont=sf]{subfig}
\usepackage{textcomp}
\usepackage[table]{xcolor}
\usepackage{stfloats}
\usepackage{url}
\usepackage{verbatim}
\usepackage{graphicx}
\usepackage{float}
\usepackage{balance}
\usepackage{algpseudocode}
\usepackage{hyperref}
\pdfpageattr{/Group <</S /Transparency /I true /CS /DeviceRGB>>}

\usepackage{eso-pic}
\newcommand{\arxivnotice}{This work has been submitted to the IEEE for possible publication.
  Copyright may be transferred without notice, after which this version may no longer be accessible.}
\AddToShipoutPictureFG{\AtPageUpperLeft{\put(0,-0.25in){\makebox[\paperwidth][c]{%
  \parbox{\textwidth}{\centering\scriptsize\arxivnotice}}}}}

\DeclareMathOperator{\explin}{explin}

\begin{document}

\title{Collision Rendering: GPU-Accelerated Trajectory Planning on Sparse Voxel Radiance Fields with Risk-Bounded Collision Checking}

\author{Byoungkwon Yoon$^{1}$, Han Gyeom Lee$^{1}$, Hojun Lee$^{1}$, Yuseop Sim$^{1}$, Daniel Sungwoo Bae$^{1}$, and \mbox{Martin Byung-Guk Jun$^{1}$}%
\thanks{Corresponding author: Martin Byung-Guk Jun}%
\thanks{$^{1}$Byoungkwon Yoon, Han Gyeom Lee, Hojun Lee, Yuseop Sim, Daniel Sungwoo Bae, and Martin Byung-Guk Jun are with School of Mechanical Engineering, Purdue University, West Lafayette, IN 47906, USA
{\tt\small \{yoon358, lee3034, lee1764, sim46, bae130, mbgjun\}@purdue.edu}}%
}

\markboth{}{}

\maketitle

\begin{figure*}[t]
    \centering
    \includegraphics[width=\textwidth]{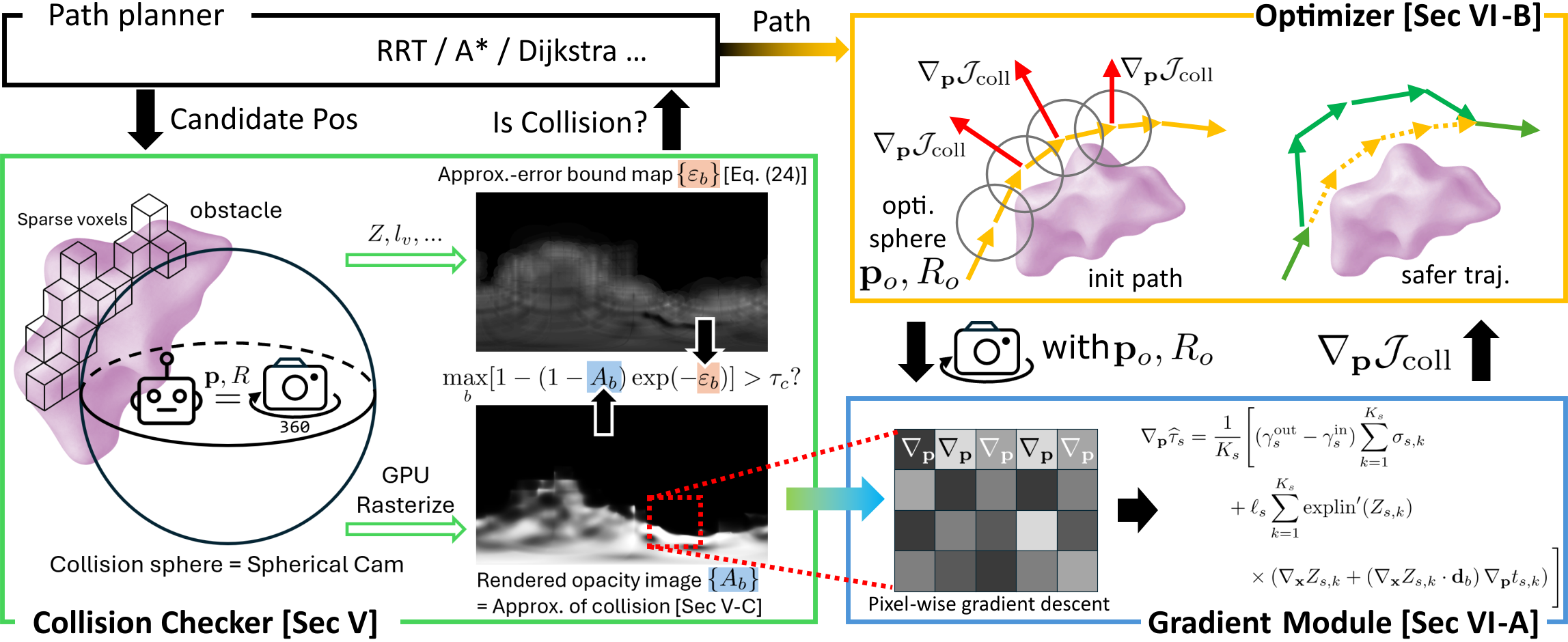}
    \caption{Overview of the proposed framework.
    Collision checker (green, Sec.~\ref{sec:collision}): the upper bound on the collision probability of a robot collision sphere is computed from the opacity image rendered by a coinciding spherical camera.
    Gradient module (blue, Sec.~\ref{sec:robot_center_alpha_gradient}): opacity images act as collision cost and are differentiated pixel-wise to obtain $\nabla_{\mathbf{p}}\mathcal{J}_{\mathrm{coll}}$.
    Optimizer (yellow, Sec.~\ref{sec:optimization process}): the collision gradients move the initial path away from obstacles.}
    \label{fig:overview}
\end{figure*}

\begin{abstract}
Although radiance fields provide detailed scene representations for robotics, they remain underexplored for trajectory planning, without a widely agreed-upon way to translate learned spatial primitives into reliable collision criteria.
To bridge this gap, we introduce Collision Rendering, the first trajectory planning framework on sparse voxel radiance fields (SVR), which supports GPU-accelerated, risk-bounded collision checking and a differentiable collision cost.
Our key insight is that the structured volumetric density of SVR allows its rendering pipeline to serve as a fast numerical approximator of collision probability.
We approximate the robot with collision spheres and place a coinciding spherical camera at each sphere.
Under a probabilistic collision model, we derive an upper bound on each sphere's maximum directional collision probability from the camera's rendering pipeline.
Evaluating this bound on the GPU yields a modular collision checker that integrates with established sampling-based planners without an auxiliary geometric map or changes to the learned representation.
We further compute a collision cost and its gradients from rendered opacities through the same pipeline, extending the framework to trajectory optimization.
Experiments demonstrate millisecond collision checking time on a consumer-grade GPU and collision-free trajectories in all queries, outperforming other radiance field-based planners.

\end{abstract}


\input{tex/introduction}

\input{tex/related_work}

\input{tex/prob_form.tex}

\input{tex/method.tex}

\input{tex/collision_bound.tex}

\input{tex/optimization.tex}

\input{tex/results}

\input{tex/discussion}

\input{tex/conclusion}

\appendices

\bibliographystyle{IEEEtran}
\bibliography{references}

\input{tex/appendix_proofs}

\end{document}

%% file: tex/introduction.tex
\section{Introduction}

Safe, collision-free trajectory planning is a fundamental requirement for robots operating in real-world environments, and it has evolved alongside the map representations used to describe these environments.
Conventional sampling-based planners, such as probabilistic roadmaps (PRMs)~\cite{kavraki1996prm} and rapidly exploring random trees (RRT)~\cite{RRT}, rely on map representations such as occupancy grids, point clouds, or polygonal meshes, which validate candidate configurations and the connections between them through occupancy queries or geometric tests~\cite{elbanhawi2014sampling}.
Alternatively, signed distance fields (SDFs) derived from geometric maps encode the signed distance to obstacle boundaries, supporting both distance-based collision checking and the obstacle-avoidance costs and gradients used in trajectory optimization~\cite{oleynikova2017voxblox}.

Although these representations support efficient motion planning, their geometric fidelity is constrained by spatial resolution and reconstruction accuracy.
The occupancy and geometric information underlying these maps is commonly derived from depth cameras or LiDAR, whose limited spatial sampling resolution can leave thin structures and fine surface details unresolved.
Even SDFs, which allow continuous distance queries, are only as accurate as the captured geometry and the field approximation.
Moreover, these representations primarily encode geometry rather than visual appearance, while preserving fine detail can incur substantial memory and computational costs.

Recent advances in radiance field methods, including NeRF~\cite{mildenhall2021nerf} and 3D Gaussian splatting (3DGS)~\cite{3dgs} variants, have opened new possibilities for dense, high-fidelity scene representation.
By modeling a scene as a continuous volumetric function optimized directly from posed images, radiance fields achieve photorealistic rendering quality while implicitly capturing geometry at arbitrary resolution.
Building on these advances, sparse voxel radiance fields (SVR)~\cite{svraster} have emerged as an alternative to 3DGS-based representations, encoding the scene in a structured voxel hierarchy rather than a set of 3D Gaussian ellipsoids.
This structured encoding resolves the ambiguous depth ordering and ill-defined volumetric density that arise from unstructured, overlapping primitives in 3DGS, while preserving the fine spatial resolution, rendering efficiency, and photorealism characteristic of radiance field methods~\cite{svraster}.

Although originally designed for photorealistic novel-view synthesis, radiance field representations inherently encode the 3D structure of a scene and have therefore been investigated for a range of robotics tasks, including simulation, real-to-sim generation, localization, and manipulation~\cite{3dgs_robot_survey}.
Their use in trajectory planning, however, remains comparatively underexplored, with only a handful of studies addressing it on radiance field representations.
These include NeRF-based approaches, such as NeRF-Nav~\cite{nerfnav} and CATNIPS~\cite{catnips}, as well as 3DGS-based approaches, such as Splat-Nav~\cite{splatnav} and Splanning~\cite{splanning}.
One reason for this limited use is that radiance fields lack a widely agreed-upon criterion for robot collision.
Although these methods demonstrate the potential of radiance fields for navigation, each evaluates collisions differently, through heuristic boundaries on density or primitive extents, robot-specific map conversions, or modifications to the underlying primitives, rather than directly on the representation as trained.
Moreover, all of them target NeRF or 3DGS, and to the best of our knowledge, trajectory planning on SVR remains unexplored despite the representational advantages outlined above.

Our key insight is that, by approximating the robot with collision spheres and modeling a spherical camera that coincides with each sphere, the rendering pipeline of SVR can be used to evaluate robot collision risk.
Under our probabilistic collision model and its assumptions, the rendering process serves as a numerical approximation of the collision probability.
Then we mathematically establish this relationship to derive a provable upper bound on the collision probability.
The bound is evaluated efficiently by leveraging the highly optimized GPU rendering pipeline of SVR, yielding a risk-bounded collision checker.
Through the same pipeline, we further compute a differentiable collision cost for gradient-based trajectory optimization.
Both components operate directly on the SVR, requiring neither conversion to another map representation nor modifications to the trained representation or heuristic collision boundaries.
Our method thus combines a provable bound on collision risk with collision queries that take only a few milliseconds on a consumer-grade GPU.

Beyond this theoretical bound, we experimentally demonstrate that our method outperforms other radiance field-based trajectory planners in simulated and real-world environments. Our contributions, illustrated in Fig.~\ref{fig:overview}, are as follows:
\begin{itemize}
    \item A modular, GPU-accelerated collision checker that operates directly on a trained sparse voxel radiance field and provides a provable upper bound on the per-direction collision probability, usable by any configuration-based planner.
    \item A gradient module that differentiates a rendered collision cost with respect to robot configurations through the same GPU rendering pipeline, which can be combined with arbitrary task costs in gradient-based trajectory optimization.
    \item A trajectory planning framework built on these two modules, combining sampling-based search with gradient-based refinement, together with experimental validation in simulated and real-world environments.
\end{itemize}

%% file: tex/related_work.tex
\section{Related Work}
We first review planning methods built on radiance field representations, followed by conventional map representations and their associated planners.

\subsection{Planning on Radiance Fields}
Radiance field methods have been applied to a range of robotic tasks, including manipulation, simulation environment construction, and SLAM~\cite{3dgs_robot_survey}, but their use for robot trajectory planning remains limited.
Most planning-related work in this area focuses on next-best-view selection for active mapping and on reaching the selected viewpoint, treating collision avoidance as a secondary concern handled through simplified heuristics rather than formal safety guarantees~\cite{li2025activesplat,jin2025activegs,jin2024gs,chen2025activegamer}.
A smaller body of work addresses trajectory planning as the primary objective: NeRF-Nav~\cite{nerfnav}, CATNIPS~\cite{catnips}, Splat-Nav~\cite{splatnav}, and Splanning~\cite{splanning}.
All four derive collision information from the density encoded in the radiance field, differing mainly in how rigorously this information is related to the true collision risk and how the resulting constraint is enforced during planning.

\textbf{NeRF-Nav.} NeRF-Nav was among the first efforts to adapt trajectory optimization to NeRF-represented environments.
It interprets the density output of a pre-trained NeRF at a point as the collision score at that point.
The robot body is approximated as a discrete set of sample points, and a trajectory is optimized to jointly minimize integrated density and control effort, using the differential flatness of quadrotor dynamics to keep the resulting trajectories dynamically feasible.
However, the density is used only as a heuristic penalty rather than a calibrated probability, so the method offers no formal safety guarantee.

\textbf{CATNIPS.} CATNIPS instead establishes a rigorous probabilistic link between NeRF density and collision risk by reinterpreting the NeRF as a Poisson point process, which allows collision probability to be derived analytically from the underlying trained scene rather than treated as a heuristic.
From this formulation, the authors construct a voxelized Probabilistically Unsafe Robot Region (PURR) that marks all robot positions violating a user-specified risk threshold, and plan through the resulting free space using an A*-seeded search refined into smooth, dynamically feasible B\'ezier trajectories via convex optimization.
However, generating the PURR requires convolving the NeRF density field with a collision-checking kernel that is parameterized by the robot's geometric shape.
As a result, any change to the robot's shape requires recomputing the entire PURR from scratch, coupling the map representation tightly to a specific robot geometry.
While these NeRF-based approaches introduce collision metrics tailored to the radiance field itself, they inherit the representation's underlying limitations, including slow training and rendering, susceptibility to floating artifacts, and limited geometric fidelity.

\textbf{Splat-Nav.} Splat-Nav targets 3DGS scenes rather than NeRFs, exploiting the fact that each Gaussian primitive defines an ellipsoidal confidence region.
Safe flight corridors are constructed through the scene using closed-form ellipsoid-ellipsoid intersection tests, and a smooth B\'ezier trajectory is then fitted within the resulting corridor by solving a quadratic program.
However, because a Gaussian primitive has no explicit boundary, Splat-Nav adopts a heuristic $1\sigma$ ellipsoid as each primitive's collision boundary, a value chosen empirically rather than derived from a formal probability bound.

\textbf{Splanning.} Splanning revisits the probabilistic view of collision risk for 3DGS scenes, introducing a normalized reformulation of the Gaussian splat representation under which the integral of density over the robot's occupied volume yields a collision probability.
Targeting articulated manipulators, Splanning overapproximates the robot's swept volume over a planning horizon using a spherical forward occupancy representation derived from reachability analysis, and enforces the resulting chance constraint as a hard constraint within a receding-horizon trajectory optimizer.
However, Splanning relies on normalized 3D Gaussian primitives to make the integrated density interpretable as a collision probability, and this normalization comes at the cost of degraded photometric and geometric reconstruction quality.
Furthermore, Splanning's reliance on an $L_\infty$-ball overapproximation of the integration domain, adopted to obtain a closed-form solution, introduces additional conservatism.

In contrast, our method uses the trained radiance field as is, with minimal assumptions, and derives a provable collision-probability bound directly from its rendering process; it requires neither heuristic collision boundaries, robot-specific map preprocessing, nor modifications to the representation that could degrade its photometric and geometric quality.

\subsection{Traditional Map Representations and Planners}
Conventional planning pipelines pair a geometric map with collision queries that guide path search or trajectory optimization.
Since these representations themselves are outside the scope of this paper, we only briefly review the related planners and refer readers to~\cite{mohanan2018survey} for a detailed survey of robotic motion planning.
Sampling-based planners such as PRM~\cite{kavraki1996prm} and RRT~\cite{RRT} use these queries to validate sampled configurations and local connections, and can therefore operate with a variety of map representations.
Common choices include occupancy grids~\cite{elfes1989occupancy} and their octree-based extension OctoMap~\cite{hornung2013octomap}, point clouds, and polygonal meshes, for which libraries such as FCL~\cite{pan2012fcl} provide efficient collision and distance queries; learned approaches such as SceneCollisionNet~\cite{danielczuk2021object} further predict collisions directly from point clouds.
Signed distance fields provide both clearance queries and gradients for trajectory optimization, as in CHOMP~\cite{ratliff2009chomp}.
These methods are well established and efficient, but applying them to an SVR requires first converting it into another map representation, for example an OctoMap built by thresholding the density or a surface mesh extracted by marching cubes.
We evaluate both conversions as baselines in Section~\ref{Sec: experimental results} and find that they have limited effectiveness for trajectory planning, whereas our method avoids them by deriving the collision-probability bound directly from the rendering process.

%% file: tex/prob_form.tex
\section{Problem Formulation and Preliminaries}
This section introduces the notation and background on which our method builds.
To link the collision probability of a robot collision sphere with the rendering pipeline of a coinciding spherical camera, we first summarize the SVR representation of SVRaster~\cite{svraster} and its rendering process.
We then define the collision probability in a density field, which is the quantity our method bounds.
Finally, we present the spherical camera model used to relate rendered opacity to this collision probability, after which Section~\ref{sec:method_overview} outlines the proposed framework.

\subsection{Notation}
We group the notation by its role below.

\subsubsection{General Conventions}
Bold symbols denote vectors, and $x_i$ denotes the $i$th component of $\mathbf{x}$.
$\mathbf{e}_i$ denotes the standard basis vector in three dimensions.
The unit sphere and its geodesic angular distance are denoted by $\mathbb{S}^2$ and $d_{\mathbb{S}^2}$, respectively.
A hat denotes a numerical approximation, and a superscript $\star$ denotes an optimized or maximizing quantity.
Indices and parameters reused across derivations are defined locally.

\subsubsection{Scene and Rendering Quantities}
The raw and activated density fields are denoted by $Z(\mathbf{x})$ and $\sigma(\mathbf{x})$.
The voxel alpha of the $i$th voxel is denoted by $\alpha_i$, rendered pixel opacity by $A$, and transmittance by $T$.

\subsubsection{Collision Quantities}
A collision sphere has center $\mathbf{p}$ and radius $R$, and $\mathbf{d}\in\mathbb{S}^2$ denotes a ray direction.
The quantities $P_{\mathrm{coll}}(\mathbf{p},\mathbf{d};R)$ and $P_{\mathrm{coll}}(\mathbf{p};R)$ denote directional and sphere-wise collision probabilities, respectively.

\subsubsection{Robot and Trajectory Quantities}
The robot configuration at waypoint $n$ is $\mathbf{q}_{n}$, and $\mathbf{q}_{0:N_{\mathrm{wp}}}$ denotes the trajectory with start and goal configurations $\mathbf{q}_s$ and $\mathbf{q}_g$.

\subsubsection{Terminology}
SVRaster~\cite{svraster} refers to the sparse-voxel-based novel-view-synthesis method, and SVR refers to its radiance field representation.

\subsection{Background: Sparse Voxel Radiance Fields}
\textbf{Scene Representation:} SVRaster~\cite{svraster} uses voxels of various sizes that are allocated in an octree to represent the scene.
Here, we summarize the characteristics that are relevant to our proposed method.
Some of the symbols are altered from the original paper's for consistency with the rest of the paper.
For details on the representation and training, please refer to the original paper.

The whole octree scene is defined by world size $w_s \in \mathbb{R}$, world center $\mathbf{w}_c \in \mathbb{R}^3$, and maximum octree level $L$.
Each voxel is defined by a triplet voxel index $\mathbf{i} = (i, j, k)$ and an octree level $l \in [1, \cdots, L]$.
Voxel size $v_s$ is defined as $v_s = w_s \cdot 2^{-l}$.
Each voxel keeps a set of spherical harmonics (SH) coefficients $\mathbf{v}_{\text{sh}}$ for view-dependent voxel color and eight raw density values $Z$.
Each raw density $Z$ corresponds to the eight corners of the voxel and is trilinearly interpolated inside the voxel.
Adjacent voxels share the same corner values, which results in a continuous density field.

\begin{figure*}[t]
    \centering
    \includegraphics[width=\textwidth]{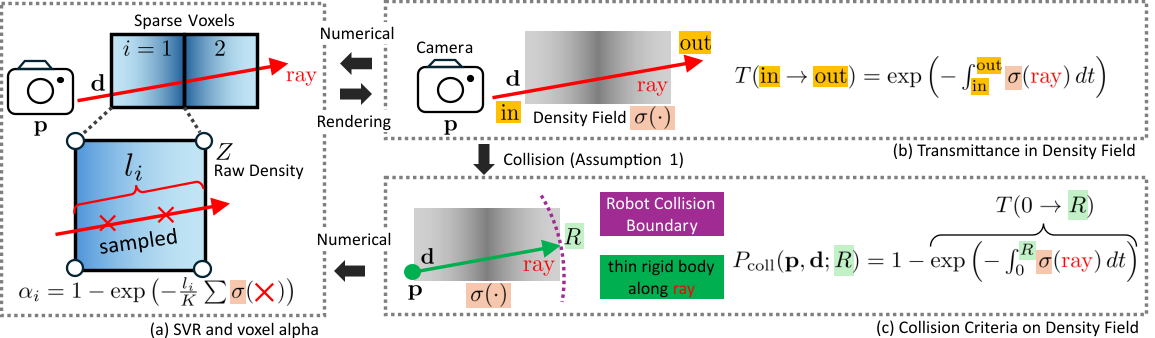}
    \caption{Example of rendering, sparse voxel field, and transmittance-collision assumption in a two-voxel scene.}
    \label{fig:transmittance}
\end{figure*}

\textbf{Rendering:} Fig.~\ref{fig:transmittance}(a) depicts the rendering process of SVRaster.
It renders each pixel through alpha compositing, as in NeRF~\cite{mildenhall2021nerf} and 3DGS~\cite{3dgs}.
The renderer first finds all voxels intersected by each pixel-ray.
Let $N$ denote the number of intersecting voxels, indexed in front-to-back order.
The pixel color $C$ is computed as
\begin{equation}
C = \sum_{i=1}^{N} T_i^\text{v} \alpha_i c_i,
\qquad T_i^\text{v} = \prod_{j=1}^{i-1} (1-\alpha_j),
\label{eq:orig_color}
\end{equation}
where $\alpha_i$ is the voxel alpha and $c_i$ is the view-dependent color of the $i$th voxel.
The $i$th voxel transmittance $T_i^\text{v}$ accounts for attenuation by preceding voxels, with $T_1^\text{v}=1$.
The voxel alpha is computed through numerical integration along its ray--voxel intersection segment:
\begin{equation}
\label{eq:orig_alpha}
\alpha_i = 1 - \exp\left(-\frac{\ell_i}{K}\sum_{k=1}^{K}\sigma(\mathbf{x}_{ik})\right),
\end{equation}
where $\ell_i$ is the segment length and $\mathbf{x}_{ik}$ is the position of the $k$th sample in voxel $i$.
The activated density at position $\mathbf{x}$ is
\begin{equation}
    \sigma(\mathbf{x}) = \explin\!\left(Z(\mathbf{x})\right),
\end{equation}
where $Z(\mathbf{x})$ is the trilinear interpolation of the eight raw corner-density values of the voxel containing $\mathbf{x}$.
The activation function $\explin$ is defined as
\begin{equation}
\label{eq:explin}
\operatorname{explin}(x)=
\begin{cases}
x, & \text{if } x > 1.1, \\[4pt]
\exp\left(\dfrac{x}{1.1}-1+\ln 1.1\right), & \text{otherwise}.
\end{cases}
\end{equation}
The rendered pixel opacity $A$, which combines the contributions of all intersected voxels, is
\begin{equation}
    A = \sum_{i=1}^{N} T_i^\text{v}\alpha_i
      = 1-\prod_{i=1}^{N}(1-\alpha_i).
    \label{eq:rendered_pixel_opacity}
\end{equation}

\textbf{Radiance and Density Fields:} SVRaster thus encodes the scene as a radiance field with two components.
The color component, given by the SH coefficients, determines the view-dependent voxel color $c_i$ and hence the pixel color $C$ in \eqref{eq:orig_color}.
The density field component $\sigma(\mathbf{x})$ determines the voxel alpha $\alpha_i$ and hence the rendered pixel opacity $A$ in \eqref{eq:rendered_pixel_opacity}.
The scene geometry is encoded implicitly in this density field, which prior planners on radiance fields likewise use to reason about collisions~\cite{nerfnav,catnips,splanning}.
Since collision checking and trajectory optimization depend on geometry, we work with the density field in the remainder of this paper, while the color component is not used.

\textbf{Pruning and Subdivision:}
During training, the octree nodes are pruned and subdivided.
Voxel nodes whose maximum blending weight $T_i^\text{v}\alpha_i$ is less than a pruning threshold are removed.
For subdivision, the top $n$ voxels with the largest training-loss gradients are subdivided and replaced by eight child nodes.
During subdivision, trilinear interpolation of the parent's corner values is used to compute the corner values of the child voxels.

\subsection{Collision Criteria on Density Field}
Similar to prior studies that relate collision probability to density fields in radiance fields~\cite{catnips,splanning,nerfnav}, we interpret transmittance in terms of collision probability.
In a density field, the transmittance $T(t_a \rightarrow t_b)$ is defined as the probability that a ray travels from distance $t_a$ to $t_b$ without interacting with any particle~\cite{tagliasacchi2022volume}.
Note that the voxel-wise transmittance $T_i^\text{v}$ in \eqref{eq:orig_color} is the same quantity in discrete form, and is included for completeness of the rendering process.
In the remainder of this paper, transmittance refers to the continuous form $T(t_a \rightarrow t_b)$.
Transmittance in SVR is depicted in Fig.~\ref{fig:transmittance}(b).
Building on this definition, we make the following assumption that connects the rendering pipeline with collision probability.

\begin{assumption}[Transmittance-collision]
\label{assump:transmittance_collision}
$T(t_a \rightarrow t_b)$ can be interpreted as the probability that an infinitesimally thin rigid body segment extending from $t_a$ to $t_b$ along the ray does not collide.
\end{assumption}

Fig.~\ref{fig:transmittance}(c) depicts the assumption.
Consider a collision sphere with center $\mathbf{p}$ and radius $R$.
For a unit direction $\mathbf{d} \in \mathbb{S}^2$,
the corresponding radial ray segment is defined as
\begin{equation}
    \mathbf{r}_{\mathbf{p},\mathbf{d}}(t)
    =
    \mathbf{p} + t\mathbf{d},
    \qquad
    t \in [0,R].
\end{equation}
Using the transmittance-collision assumption, the collision probability along direction $\mathbf{d}$ is defined as the
complement of the transmittance from the sphere center to its boundary:
\begin{equation}
    P_{\mathrm{coll}}(\mathbf{p},\mathbf{d};R)
    =
    1 - T_{\mathbf{p},\mathbf{d}}(0 \rightarrow R).
    \label{eq:collision_prob_ray}
\end{equation}
\begin{definition}[Maximum directional collision probability]
\label{def:max_dir_coll_prob}
The collision probability of a collision sphere is the maximum of its directional collision probability over all directions:
\begin{equation}
    P_{\mathrm{coll}}(\mathbf{p};R)
    =
    \max_{\mathbf{d} \in \mathbb{S}^2}
    \left[
        1 - T_{\mathbf{p},\mathbf{d}}(0 \rightarrow R)
    \right].
    \label{eq:spherical_collision_probability}
\end{equation}
The sphere is declared to be in collision if $P_{\mathrm{coll}}(\mathbf{p};R)\geq\tau_c$.
\end{definition}

Since $P_{\mathrm{coll}}(\mathbf{p};R)$ is the largest collision probability among all radial segments, keeping its upper bound below $\tau_c$ bounds the collision risk of every segment, i.e., it imposes an individual chance constraint on each radial segment~\cite{ono2008iterative}, analogous to max-aggregation of occupancy probabilities in safe navigation~\cite{lunenburg2018representation}.
Theorem~\ref{thm:sphere_max_upper_bound} certifies this bound for all directions in $\mathbb{S}^2$, which is the sense in which the collision checks of the proposed planner are risk-bounded.
Splanning~\cite{splanning} adopts the same ray-from-the-center model but bounds the fraction of uniformly sampled directions whose collision probability exceeds the threshold, using a volume integral of a normalized density; in contrast, \eqref{eq:spherical_collision_probability} requires every direction to satisfy the threshold, which makes the proposed criterion the strictest instance of this model.
Using the volume-rendering definition of transmittance~\cite{max2010local},
\begin{equation}
    \label{eq:transmittance}
    T_{\mathbf{p},\mathbf{d}}(0 \rightarrow R)
    =
    \exp\left(
        -\int_{0}^{R}
        \sigma\left(\mathbf{r}_{\mathbf{p},\mathbf{d}}(t)\right)
        \, \mathrm{d}t
    \right),
\end{equation}
the collision probability becomes
\begin{equation}
    \label{eq:coll_prob_def}
    P_{\mathrm{coll}}(\mathbf{p};R)
    =
    \max_{\mathbf{d} \in \mathbb{S}^2}
    \left[
    1 -
        \exp\left(
            -\int_{0}^{R}
            \sigma\left(\mathbf{r}_{\mathbf{p},\mathbf{d}}(t)\right)
            \, \mathrm{d}t
        \right)
    \right].
\end{equation}

\subsection{Spherical Camera Model and Rendering}
\label{subsec:cam_model}
To relate the collision probability of a robot collision sphere to rendering, we need a camera model that coincides with the sphere.
Since the original SVRaster supports only planar (pinhole) image planes, we introduce a spherical camera model for SVR and the corresponding rendering procedure.
The resulting spherical image is related to the robot collision sphere in the following sections.

For image pixel coordinate $(x,y)$, the camera-space angular coordinates with image resolution of $W \times H$ are:
\begin{equation}
    \theta = 2\pi\frac{x+0.5-c_x}{W}, \quad \phi=\pi\frac{y+0.5-c_y}{H}.
\end{equation}
The corresponding camera-space direction is
\begin{equation}
    \mathbf{d}_{\mathrm{cam}} =
    \begin{bmatrix}
        \sin\theta \cos\phi \\
        \sin\phi \\
        \cos\theta \cos\phi
    \end{bmatrix}.
\end{equation}
The direction is transformed into world coordinates and normalized:
\begin{equation}
    \label{eq:pixel-ray}
    \mathbf{d} = \frac{\mathbf{R}_{\mathrm{c2w}}\mathbf{d}_{\mathrm{cam}}}{\lVert\mathbf{R}_{\mathrm{c2w}}\mathbf{d}_{\mathrm{cam}}\rVert_2}.
\end{equation}
The world-space pixel-ray is therefore
\begin{equation}
    \mathbf{r}(t) = \mathbf{p} + t\mathbf{d}.
\end{equation}

Given the pixel ray, the remaining rendering steps follow the original SVRaster on the GPU.
For each pixel ray, ray--AABB tests against the candidate voxels determine the intersected voxels and their ray--voxel segments.
Details of the pixel rendering process can be found in~\cite{svraster}.
The resulting pixel opacity is computed using \eqref{eq:rendered_pixel_opacity}; its relationship to collision probability is established in Section~\ref{subsec:rendering_approximation}.

%% file: tex/method.tex
\section{Method Overview}
\label{sec:method_overview}

\renewcommand{\algorithmicrequire}{\textbf{Input:}}
\renewcommand{\algorithmicensure}{\textbf{Output:}}

\begin{algorithm}[!t]
\caption{Path Planner}
\label{alg:Path_planner}
\footnotesize
\begin{algorithmic}[1]
\Require SVR $\mathcal{M}$, spherical opacity renderer $\mathcal{R}$, and path planner $\mathcal{P}$
\Require Start and goal configurations $\mathbf{q}_{s},\mathbf{q}_{g}$
\Require Collision spheres $\{(\mathbf{p}_{m}(\mathbf{q}),R_m)\}_{m=1}^{M}$ and threshold $\tau_c$
\Ensure Initial path $\mathbf{q}_{0:N_{\mathrm{wp}}}^{0}$

\Function{IsSafe}{$\mathbf{q}$}
    \For{$m \gets 1$ to $M$}
        \State $\{A_b\}_{b=1}^{B} \gets \mathcal{R}(\mathbf{p}_m,R_m,\mathcal{M})$
        \Comment{render opacity, Sec.~\ref{subsec:cam_model}}
        \If{$\max_{b=1,\ldots,B} A_b \geq \tau_c$}
            \State \Return \textbf{false}
        \EndIf
        \State $A^{\mathrm{ub}} \gets \Call{UpperBound}{\{A_b\}_{b=1}^B}$
        \Comment{Sec.~\ref{sec:collision}, Eq.~\eqref{eq:rendered_collision_upper}}
        \If{$A^{\mathrm{ub}} \geq \tau_c$}
            \State \Return \textbf{false}
        \EndIf
    \EndFor
    \State \Return \textbf{true}
\EndFunction

\State $\mathbf{q}_{0:N_{\mathrm{wp}}}^{0} \gets
    \Call{Plan}{\mathcal{P},\mathbf{q}_s,\mathbf{q}_g,\textsc{IsSafe}}$
\If{$\mathbf{q}_{0:N_{\mathrm{wp}}}^{0}=\textsc{Failure}$}
    \State \Return \textsc{Failure}
\EndIf
\State \Return $\mathbf{q}_{0:N_{\mathrm{wp}}}^{0}$
\end{algorithmic}
\end{algorithm}

Our method provides two modules for trajectory planning in scenes represented by SVR: (1) a collision checker that computes an upper bound on the collision probability of a given collision sphere through the rendering pipeline, and (2) a gradient evaluator that computes the spatial gradient of a collision cost with respect to the sphere's center.
The collision checker determines whether a robot configuration satisfies a prescribed risk threshold, while the gradient evaluator guides trajectory updates to reduce the collision cost.
Our trajectory planning framework integrates these modules by combining a sampling-based path planner with a gradient-based trajectory optimizer, as summarized in Algorithms~\ref{alg:Path_planner} and~\ref{alg:Traj_Opti}.
The following sections present the mathematical derivations supporting collision checking and gradient computation.

We first describe the collision checker, presented in Algorithm~\ref{alg:Path_planner}.
We represent the robot's collision geometry using $M$ spheres, where the $m$th sphere has center $\mathbf{p}_m(\mathbf{q})$ at robot configuration $\mathbf{q}$ and radius $R_m$.
We assume that the union of these spheres fully contains the robot's geometry.
For each sphere, a spherical camera that coincides with it renders an opacity image with resolution $W\times H$ and pixel opacities $\{A_b\}_{b=1}^{B}$, where $B=WH$.
A configuration is declared safe only if, for every sphere, both the maximum pixel opacity and the upper bound computed from the opacity image remain below the threshold $\tau_c$.
The collision checker certifies individual configurations and is independent of the planner; how edges between configurations are validated is determined by the planner.
In our implementation, edges and segments between consecutive waypoints are validated by evaluating \textsc{IsSafe($\mathbf{q}$)} at configurations interpolated with a fixed step.

We next describe the trajectory optimizer, summarized in Algorithm~\ref{alg:Traj_Opti}.
Starting from the path returned by the sampling-based planner, we refine the intermediate robot configurations while keeping the start and goal fixed.
At each iteration, we render an opacity image of a single optimization sphere at each waypoint and aggregate its pixel values into a collision cost using an image-cost aggregator $\rho$.
We differentiate this cost with respect to the sphere center and map the resulting spatial gradient to robot configuration space through the Jacobian of the sphere center.
These gradients are combined with those of a task cost, such as a trajectory smoothness cost, to compute a candidate gradient-descent update.
The update is accepted only if the collision checker verifies that every collision sphere satisfies the prescribed risk threshold at all sampled configurations along the candidate trajectory; otherwise, the previous trajectory is retained.

\begin{algorithm}[!t]
\caption{Trajectory Optimizer}
\label{alg:Traj_Opti}
\footnotesize
\begin{algorithmic}[1]
\Require SVR $\mathcal{M}$, spherical opacity renderer $\mathcal{R}$
\Require Initial path $\mathbf{q}_{0:N_{\mathrm{wp}}}^0$
\Require Collision spheres $\{(\mathbf{p}_{m}(\mathbf{q}),R_m)\}_{m=1}^{M}$, optimization sphere $(\mathbf{p}_{\mathrm{o}}(\mathbf{q}),R_{\mathrm{o}})$, and threshold $\tau_c$
\Require Task cost $\mathcal{J}_{\mathrm{task}}$, image-cost aggregator $\rho$, weights $\lambda_{\mathrm{task}},\lambda_{\mathrm{coll}}$
\Require Iteration limit $K$, tolerance $\epsilon$, and decreasing step sizes $\{\eta_k\}_{k=0}^{K-1}$
\Ensure Optimized trajectory $\mathbf{q}_{0:N_{\mathrm{wp}}}$

\For{$k \gets 0$ to $K-1$}
    \State $(\mathcal{J}_{\mathrm{task}},\mathbf{g}_{\mathrm{task}})
        \gets \Call{TaskCostGrad}{\mathbf{q}_{0:N_{\mathrm{wp}}}^{k}}$
    \State $\mathcal{J}_{\mathrm{coll}} \gets 0$;
        $\mathbf{g}_{\mathrm{coll}} \gets \mathbf{0}$
    \For{$n \gets 1$ to $N_{\mathrm{wp}}-1$}
        \State $I_{n} \gets \mathcal{R}(\mathbf{p}_{o}(\mathbf{q}_n^k),R_{\mathrm{o}},\mathcal{M})$
        \Comment{render opacity image}
        \State $\mathcal{J}_{\mathrm{coll}} \gets
            \mathcal{J}_{\mathrm{coll}}+\rho(I_{n})$
        \Comment{update collision cost}
        \State $\mathbf{g}_{\mathrm{coll},n} \gets
            J_{\mathbf{p}_{\mathrm{o}}}(\mathbf{q}_{n}^{k})^{\mathsf T}
            \nabla_{\mathbf{p}_{n}}\rho(I_{n})$
        \Comment{Sec.~\ref{sec:robot_center_alpha_gradient}}
    \EndFor
    \State $\mathcal{J} \gets \lambda_{\mathrm{task}}\mathcal{J}_{\mathrm{task}}
        +\lambda_{\mathrm{coll}}\mathcal{J}_{\mathrm{coll}}$
    \State $\mathbf{g} \gets \lambda_{\mathrm{task}}\mathbf{g}_{\mathrm{task}}
        +\lambda_{\mathrm{coll}}\mathbf{g}_{\mathrm{coll}}$
    \If{$\lVert\mathbf{g}\rVert_2 \leq \epsilon$}
        \State \Return $\mathbf{q}_{0:N_{\mathrm{wp}}}^{k}$
    \EndIf
    \State $\mathbf{q}_{0:N_{\mathrm{wp}}}^{\mathrm{cand}} \gets \mathbf{q}_{0:N_{\mathrm{wp}}}^{k}$
    \For{$n \gets 1$ to $N_{\mathrm{wp}}-1$}
        \State $\mathbf{q}_{n}^{\mathrm{cand}} \gets
            \mathbf{q}_{n}^{k}-\eta_k\mathbf{g}_{n}$
    \EndFor
    \If{$\Call{IsSafe}{\mathbf{q}}$ for all
        $\mathbf{q}\in\Call{Samples}{\mathbf{q}_{0:N_{\mathrm{wp}}}^{\mathrm{cand}}}$}
        \State $\mathbf{q}_{0:N_{\mathrm{wp}}}^{k+1} \gets \mathbf{q}_{0:N_{\mathrm{wp}}}^{\mathrm{cand}}$
    \Else
        \State $\mathbf{q}_{0:N_{\mathrm{wp}}}^{k+1} \gets \mathbf{q}_{0:N_{\mathrm{wp}}}^{k}$
        \Comment{reject unsafe update}
    \EndIf
\EndFor
\State \Return $\mathbf{q}_{0:N_{\mathrm{wp}}}^{K}$
\end{algorithmic}
\end{algorithm}

%% file: tex/collision_bound.tex
\section{Collision Probability from Rendering} \label{sec:collision}

We connect the rendering pipeline with collision probability in Section~\ref{subsec:rendering_approximation} and state the sphere-wise collision bound in Section~\ref{subsec:sphere_bound}; supporting proofs are provided in Appendix~\ref{app:collision_proofs}.

Consider a collision sphere centered at $\mathbf{p}$ with radius $R$.
Since the collision check in Algorithm~\ref{alg:Path_planner} is applied to each of the $M$ collision spheres independently, we consider a single collision sphere from now on without loss of generality.

\subsection{Direct Integration on Sparse Voxel Radiance Fields}
Before proving the upper bound, we first examine the exact integral in \eqref{eq:coll_prob_def} on the SVR.
Recall from \eqref{eq:explin} that the trainable variable determining the voxel alpha is the raw density value $Z(\mathbf{x})$, with activated density $\sigma(\mathbf{x}) = \operatorname{explin}(Z(\mathbf{x}))$.
Also, the value of $Z(\mathbf{x})$ inside the voxel is trilinearly interpolated.
As a result, within a single voxel, the exact integral along the ray $\mathbf{r}_{\mathbf{p},\mathbf{d}}(t)$ over its ray--voxel segment $s$ is
\begin{equation}
    \int_{t_s^{\mathrm{in}}}^{t_s^{\mathrm{out}}} \operatorname{explin}\left(c_{s,3}t^3 + c_{s,2}t^2 + c_{s,1}t + c_{s,0}\right) \mathrm{d}t,
    \label{eq:exact_ray_integral}
\end{equation}
where $[t_s^{\mathrm{in}},t_s^{\mathrm{out}}]\subseteq[0,R]$ is the interval of the segment and $c_{s,3},\ldots,c_{s,0}$ are coefficients determined by the trilinear interpolation inside the voxel.
On the exponential branch of $\operatorname{explin}$, a quadratic or cubic raw-density polynomial along a ray--voxel segment yields an integrand with no elementary antiderivative, as follows from Liouville's theorem on elementary integration~\cite{rosenlicht1972integration}.
Risk-bounded collision checking must support repeated planning queries while certifying that the exact collision probability remains below the prescribed risk threshold.
This motivates an upper bound that explicitly bounds both approximation errors while reusing the renderer's optimized evaluation pipeline.

\subsection{Assumptions}
In this subsection, we present the assumptions required for the proof.
\begin{assumption}[Scene continuity]
\label{assump:scene_continuity}
Adjacent voxels inside the sphere share corner values, so the raw density field $Z(\cdot)$ and activated density field $\sigma(\cdot)$ are continuous across voxel boundaries.
\end{assumption}

In a trained SVR, the density field can be discontinuous where adjacent voxels belong to different octree levels or where pruning leaves empty space.
We therefore evaluate collisions on a local copy of the voxels enclosed by the sphere, in which all enclosed voxels are subdivided to the maximum octree level $L$, which removes discontinuities between octree levels~\cite{svraster}, and empty regions are filled with voxels of negligible density whose corners shared with existing voxels retain their trained values.
This local density field satisfies the assumption, while the pre-trained representation itself remains unchanged.

\begin{assumption}[Angular cover]
\label{assump:angular_cover}
The sphere is covered by $B$ angular bins $\{\mathcal{B}_b\}_{b=1}^{B}$.
Each bin has center direction $\mathbf{d}_b$ and angular radius $\delta_b$:
\begin{equation}
    \label{eq:angular_bin_radius}
    \mathcal{B}_b \subseteq \{\mathbf{d} \in \mathbb{S}^2: d_{\mathbb{S}^2}(\mathbf{d}, \mathbf{d}_b) \leq \delta_b\}
\end{equation}
Together, the angular bins $\{\mathcal{B}_b\}_{b=1}^{B}$ cover the sphere:
    \[
        \mathbb{S}^{2}
        \subseteq
        \bigcup_{b=1}^{B} \mathcal{B}_b.
    \]
\end{assumption}

\begin{remark}
The derivative of $\operatorname{explin}$ satisfies
\[
0<
\operatorname{explin}'(x)
\le 1
\qquad
\forall x\in\mathbb{R}.
\]
Therefore, by the chain rule,
\begin{equation}
    \|\nabla\sigma(\mathbf{x})\|_{2}
    =
    \left|
    \operatorname{explin}'(Z(\mathbf{x}))
    \right|
    \|\nabla Z(\mathbf{x})\|_{2}
    \le
    \|\nabla Z(\mathbf{x})\|_{2}.
    \label{eq:explin_gradient_bound}
\end{equation}
Hence, any valid Lipschitz constant of the raw density field $Z(\cdot)$ is also a valid Lipschitz constant of the density field $\sigma$.
\end{remark}

For a voxel $v$ with side length
$l_v$ and raw corner density
$Z^{(v)}_{abc}$, $a,b,c\in\{0,1\}$, define
\begin{equation}
\begin{aligned}
    L_{\text{vox},v}
    =
    \max_{a,b,c\in\{0,1\}}
    \Bigg[
    &
    \left(
    \frac{
    Z^{(v)}_{1bc}-Z^{(v)}_{0bc}
    }{l_v}
    \right)^{2}
    +
    \left(
    \frac{
    Z^{(v)}_{a1c}-Z^{(v)}_{a0c}
    }{l_v}
    \right)^{2}
    \\[-2pt]
    &+
    \left(
    \frac{
    Z^{(v)}_{ab1}-Z^{(v)}_{ab0}
    }{l_v}
    \right)^{2}
    \Bigg]^{1/2}.
\end{aligned}
\label{eq:voxel_lipschitz}
\end{equation}
For a trilinearly interpolated voxel, \eqref{eq:voxel_lipschitz} is the maximum gradient norm inside the voxel.

\subsection{Rendering and Collision-Probability Approximation}
\label{subsec:rendering_approximation}
Our key idea is that the rendered opacity image of a spherical camera whose sphere coincides with the robot collision sphere acts as a numerical approximation of the collision probability.
We can render the opacity image with the spherical camera model defined in Section~\ref{subsec:cam_model} with the same center and radius.
Then, the rendered opacity image is an approximation of the collision probability.
The opacity image rendering approximates the collision probability in two ways.

First, the rendering pipeline samples density along a pixel-center ray and this numerically approximates the integrated density in \eqref{eq:coll_prob_def}.
Let $\tau(\mathbf{d})$ denote this exact integral and $\widehat{\tau}(\mathbf{d})$ its rendering approximation.
Let $\mathcal{S}(\mathbf{d})$ denote the ray--voxel intersection segments within radius $R$ along direction $\mathbf{d}$.
This set also includes voxels that are only partially enclosed by the sphere, namely the voxel containing the sphere center and voxels crossing its boundary; for these, only the portion of the segment within $[0,R]$ is used.
For each segment $s$, let $\ell_s$ denote its length and let $\mathbf{x}_{s,k}$ denote its $K_s$ evenly spaced sample positions.
Equations~\eqref{eq:orig_alpha} and~\eqref{eq:rendered_pixel_opacity} then give
\begin{equation}
\begin{aligned}
    \widehat{\tau}_s
    &= \frac{\ell_s}{K_s}
    \sum_{k=1}^{K_s}\sigma(\mathbf{x}_{s,k}), \\
    \widehat{\tau}(\mathbf{d})
    &= \sum_{s\in\mathcal{S}(\mathbf{d})}\widehat{\tau}_s, \\
    A(\mathbf{d})
    &= 1-\exp\bigl(-\widehat{\tau}(\mathbf{d})\bigr).
\end{aligned}
    \label{eq:pixel_segment_opacity}
\end{equation}
Here, $\widehat{\tau}_s$ is the sampled integrated density over segment $s$, and $A(\mathbf{d})$ is the rendered pixel opacity.
Since $\widehat{\tau}(\mathbf{d})$ approximates $\tau(\mathbf{d})$, $A(\mathbf{d})$ approximates the directional collision probability in \eqref{eq:collision_prob_ray}.

Second, rendering a spherical opacity image is the same as approximating the collision probability over the sphere by sampling along the pixel-center ray directions.
The rendered opacity image therefore can be interpreted as an approximation map of the directional collision probability sampled over the sphere, as depicted in Fig.~\ref{fig:overview}.

However, since we want the collision checker to be risk-bounded, an approximated value is not enough, and we need an upper bound on the collision probability, which motivates the bound derived next.
For this, in the following subsection, we prove that the collision probability in \eqref{eq:coll_prob_def} can be upper bounded using the rendered spherical opacity image.
We refer to the errors of the two approximations above as the ray and angular approximation errors, or ray and angular errors for short, and to their upper bounds as error bounds.

\begin{figure*}[t]
    \centering
    \includegraphics[width=\textwidth]{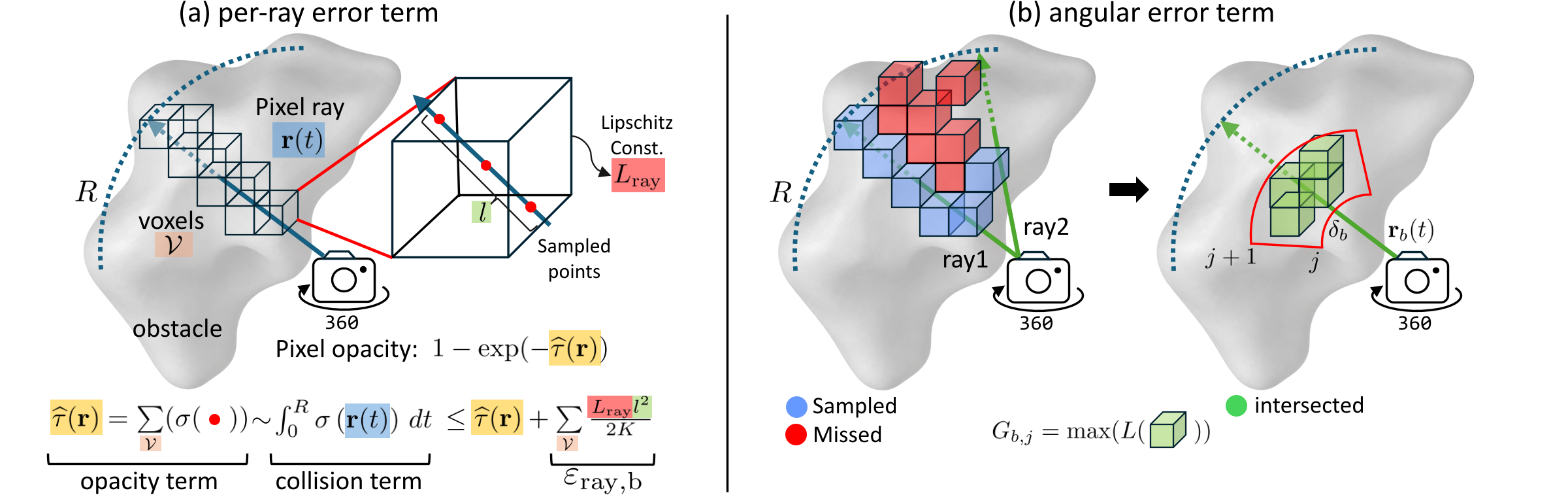}
    \caption{Error terms of the collision upper bound. (a) Per-ray error: the rendered pixel opacity sums the density at discrete samples (red dots) along the pixel ray $\mathbf{r}(t)$, and the quadrature error in each voxel is bounded using its ray Lipschitz constant $L_{\mathrm{ray}}$ and segment length. (b) Angular error: voxels between neighboring pixel rays (red) can be missed by the sampled rays (green). The angular error bound covers the density over the cone of half-angle $\delta_b$ around the bin ray $\mathbf{r}_b(t)$, using the voxels in each radial shell.}
    \label{fig:upperbound}
\end{figure*}

\subsection{Sphere-Wise Upper Bound}
\label{subsec:sphere_bound}
For each angular bin, we combine a per-ray upper bound $U_{\mathrm{ray},b}$ with an angular approximation-error bound $\varepsilon_{\mathrm{ang},b}$, then take the maximum over all bins.
Detailed definitions and the supporting bounds are provided in Appendix~\ref{app:collision_proofs}.

\begin{theorem}[Sphere-maximum upper bound]
\label{thm:sphere_max_upper_bound}
Define the upper bound for angular bin $\mathcal{B}_b$ as
\begin{equation}
    U_b
    =
    U_{\mathrm{ray},b}
    +
    \varepsilon_{\mathrm{ang},b}.
    \label{eq:angular_bin_upper}
\end{equation}
Then
\begin{equation}
    \tau_{\max}
    =
    \max_{\mathbf{d}\in\mathbb{S}^{2}}
    \tau(\mathbf{d})
    \le
    \max_{b=1,\ldots,B}
    U_b.
    \label{eq:sphere_tau_upper}
\end{equation}
Substituting $U_{\mathrm{ray},b}$ and $\varepsilon_{\mathrm{ang},b}$ from Theorem~\ref{thm:per_ray_upper_bound} and Lemma~\ref{lem:angular_deviation} in Appendix~\ref{app:collision_proofs} gives
\begin{align}
    \tau_{\max}
    \le
    \max_b
    \Bigg\{
    &
    \underbrace{
    \sum_{s\in\mathcal{S}(\mathbf{d}_b)}
    \left[
        \widehat{\tau}_s
        +
        \frac{
        L_{\mathrm{ray},s}\ell_s^{2}
        }{
        2K_s
        }
    \right]
    }_{U_{\mathrm{ray},b}}
    \nonumber\\
    &+
    \underbrace{
    2\sin\left(\frac{\delta_b}{2}\right)
    \sum_{j=0}^{N_r-1}
    G_{b,j}
    \frac{
    r_{j+1}^{2}-r_j^{2}
    }{2}
    }_{\varepsilon_{\mathrm{ang},b}}
    \Bigg\},
    \label{eq:combined_sphere_upper}
\end{align}
where $N_r$ is the total number of radial shells.
The maximum directional collision probability $P_{\mathrm{coll}}(\mathbf{p};R)$ is consequently bounded by
\begin{equation}
    P_{\mathrm{coll}}(\mathbf{p};R)
    \le
    1-
    \exp\left(
        -
        \max_b U_b
    \right).
    \label{eq:sphere_alpha_upper}
\end{equation}
\end{theorem}
\noindent The proof is provided in Appendix~\ref{app:proof_sphere_bound}.

The $B=WH$ pixels of a spherical image with resolution $W\times H$ define angular bins $\mathcal{B}_b$ containing the viewing directions within each pixel.
We use the corresponding pixel-center ray direction in \eqref{eq:pixel-ray} as the bin center $\mathbf{d}_b$.
The azimuth and vertical angular half-widths are $\pi/W$ and $\pi/(2H)$, respectively, so the uniform choice
\[
    \delta_b=\delta
    =\sqrt{\left(\frac{\pi}{W}\right)^2
        +\left(\frac{\pi}{2H}\right)^2}
\]
satisfies the angular-radius condition in \eqref{eq:angular_bin_radius} for every pixel.

Let $A_b=A(\mathbf{d}_b)$ denote the rendered pixel opacity for pixel ray direction $\mathbf{d}_b$.
By \eqref{eq:pixel_segment_opacity}, its accumulated density is $\widehat{\tau}(\mathbf{d}_b)=\sum_{s\in\mathcal{S}(\mathbf{d}_b)}\widehat{\tau}_s$.
Define the total error bound $\varepsilon_b$ for this bin as
\[
    \varepsilon_b
    =\varepsilon_{\mathrm{ray},b}
    +\varepsilon_{\mathrm{ang},b},
    \qquad
    \varepsilon_{\mathrm{ray},b}
    =\sum_{s\in\mathcal{S}(\mathbf{d}_b)}
    \frac{L_{\mathrm{ray},s}\ell_s^2}{2K_s}.
\]
Figure~\ref{fig:upperbound}(a) describes the first ray term, and (b) describes the second angular term.
The per-bin upper bound can therefore be written as $U_b=\widehat{\tau}(\mathbf{d}_b)+\varepsilon_b$.
Let $A_{\max}=\max_b A_b$ denote the maximum rendered pixel opacity, which approximates $P_{\mathrm{coll}}(\mathbf{p};R)$.
Since $A_b=1-\exp(-\widehat{\tau}(\mathbf{d}_b))$, we have $1-\exp(-U_b)=1-(1-A_b)\exp(-\varepsilon_b)$, and \eqref{eq:sphere_alpha_upper} becomes
\begin{equation}
    P_{\mathrm{coll}}(\mathbf{p};R)
    \leq A^{\mathrm{ub}}
    = \max_b\left[1-(1-A_b)\exp(-\varepsilon_b)\right].
    \label{eq:rendered_collision_upper}
\end{equation}
Because $\varepsilon_b\geq 0$, the bound satisfies $A^{\mathrm{ub}}\geq A_{\max}$.

Equation~\eqref{eq:rendered_collision_upper} supports the two-stage collision check in Algorithm~\ref{alg:Path_planner}.
Since $A^{\mathrm{ub}}$ is no smaller than $A_{\max}$, a configuration is rejected before evaluating the error bounds if $A_{\max}\geq\tau_c$ for any collision sphere.
Otherwise, $A^{\mathrm{ub}}$ is computed, and the configuration is accepted only if the bound is below the prescribed risk threshold $\tau_c$ for every collision sphere.

The upper bound in \eqref{eq:rendered_collision_upper} reuses the pixel opacities $A_b$ and requires the voxel-related quantities $L_{\mathrm{ray},s}$, $\ell_s$, and $G_{b,j}$, which can be queried or computed during rendering; the remaining coefficients are determined by fixed camera and sampling settings.
Consequently, bound evaluation can be integrated into the existing GPU rendering pipeline.
Algorithms~\ref{alg:gpu_ray_lipschitz} and~\ref{alg:gpu_angular_error} in Appendix~\ref{app:collision_proofs} summarize the ray and angular error-bound computations, respectively.

%% file: tex/optimization.tex
\section{Rendering-Based Trajectory Optimization}
Starting from the path returned by Algorithm~\ref{alg:Path_planner}, we refine the trajectory using Algorithm~\ref{alg:Traj_Opti}.
Our use of rendering gradients for trajectory optimization draws inspiration from iNeRF~\cite{yen2021inerf}, which refines uncertain camera poses through differentiable rendering.
Using the approximation established in Section~\ref{subsec:rendering_approximation}, we adopt rendered pixel opacities as a surrogate collision cost to guide robot configuration updates.
Thus, the collision cost is computed by the same spherical camera rendering of the SVR that the collision checker uses.
Only the rendered sphere differs, an optimization sphere of radius $R_{\mathrm{o}}$ instead of the robot collision spheres, so checking and optimization share one rendering pipeline and require no additional representation.
The gradient of each pixel opacity is evaluated analytically in a CUDA kernel that implements the derivative in Appendix~\ref{app:optimization_proofs}.
Since each pixel gradient depends only on the samples along its own ray, the kernel processes all pixels in parallel, in the same manner as the forward rendering.
The collision checker enforces the prescribed risk threshold using the upper bound in \eqref{eq:rendered_collision_upper}.
In the following subsections, we derive the gradient of the rendered pixel opacity with respect to the sphere center and then describe the optimization process.

\subsection{Gradient of Rendered Pixel Opacity}
\label{sec:robot_center_alpha_gradient}
Using the rendering quantities in \eqref{eq:pixel_segment_opacity}, we differentiate the pixel opacity $A_b$ with respect to the sphere center $\mathbf{p}$ while holding the ray direction $\mathbf{d}_b$ fixed.
For each segment $s\in\mathcal{S}(\mathbf{d}_b)$, let $[t_s^{\mathrm{in}},t_s^{\mathrm{out}}]\subseteq[0,R]$ be its interval, with $\ell_s=t_s^{\mathrm{out}}-t_s^{\mathrm{in}}$.
We parameterize its $K_s$ midpoint samples, indexed by $k=1,\ldots,K_s$, as
\[
\begin{gathered}
    u_{s,k}=\frac{k-1/2}{K_s},
    \qquad t_{s,k}=(1-u_{s,k})t_s^{\mathrm{in}}+u_{s,k}t_s^{\mathrm{out}}, \\
    \mathbf{x}_{s,k}=\mathbf{p}+t_{s,k}\mathbf{d}_b.
\end{gathered}
\]
Write $Z_{s,k}=Z(\mathbf{x}_{s,k})$ and $\sigma_{s,k}=\sigma(\mathbf{x}_{s,k})$, and define the endpoint and sample-distance gradients as
\[
\begin{gathered}
    \gamma_s^{\mathrm{in}}=\nabla_{\mathbf{p}}t_s^{\mathrm{in}},
    \qquad\gamma_s^{\mathrm{out}}=\nabla_{\mathbf{p}}t_s^{\mathrm{out}}, \\
    \nabla_{\mathbf{p}}t_{s,k}=(1-u_{s,k})\gamma_s^{\mathrm{in}}+u_{s,k}\gamma_s^{\mathrm{out}}.
\end{gathered}
\]
We assume that the active intersection axes are held fixed during differentiation; an endpoint at $t=0$ or $t=R$ has zero derivative.

\begin{theorem}[Sphere-center derivative of one voxel]
\label{thm:voxel_center_derivative}
    For segment $s$ and the midpoint samples defined above, the gradient of the sampled integrated density is
    \begin{equation}
        \label{eq:voxel_center_derivative}
        \begin{aligned}
        \nabla_\mathbf{p} \widehat\tau_s
        &=
        \frac{1}{K_s}
        \Bigg[
        (\gamma_s^{\mathrm{out}}-\gamma_s^{\mathrm{in}})
        \sum_{k=1}^{K_s}\sigma_{s,k}
        \\
        &\qquad
        +
        \ell_s
        \sum_{k=1}^{K_s}
        \explin'(Z_{s,k})
        \\
        &\qquad\quad {}
        \times
        \left(
        \nabla_\mathbf{x} Z_{s,k}
        +
        (\nabla_\mathbf{x} Z_{s,k}\cdot \mathbf{d}_b)\nabla_{\mathbf{p}}t_{s,k}
        \right)
        \Bigg].
        \end{aligned}
    \end{equation}
\end{theorem}

Combining these segment gradients with the pixel-opacity relation in \eqref{eq:pixel_segment_opacity} gives
\begin{equation}
    \nabla_{\mathbf{p}} A_b
    =(1-A_b)
    \sum_{s\in\mathcal{S}(\mathbf{d}_b)}
    \nabla_{\mathbf{p}}\widehat{\tau}_s.
    \label{eq:pixel_center_gradient}
\end{equation}
The proofs of Theorem~\ref{thm:voxel_center_derivative} and of this relation (Corollary~\ref{cor:pixel_opacity_gradient}) are provided in Appendices~\ref{app:proof_voxel_gradient} and~\ref{app:pixel_opacity_gradient}, respectively.

\subsection{Optimization Process}
\label{sec:optimization process}
We optimize the intermediate robot configurations while keeping the start and goal fixed.
The collision cost is computed from a single optimization sphere with center $\mathbf{p}_{\mathrm{o}}(\mathbf{q})$ and radius $R_{\mathrm{o}}$, chosen to enclose the region around the robot whose clearance should be increased, while the $M$ collision spheres are retained for collision checking.
Since the initial path waypoints already satisfy the collision criterion, rendering over the collision spheres may provide limited gradient information about nearby obstacles.
A sufficiently large $R_{\mathrm{o}}$ allows these obstacles to contribute to the optimization cost and encourages additional clearance.
For waypoint $n$, let $\mathbf{p}_{n}=\mathbf{p}_{\mathrm{o}}(\mathbf{q}_{n})$ and let $I_{n}$ denote the opacity image rendered at $\mathbf{p}_{n}$ with radius $R_{\mathrm{o}}$.
Its pixel values are $A_{n,b}$, namely $A_b$ evaluated for this sphere, and $\rho(I_{n})$ aggregates these values into a collision cost.
The trajectory optimization problem is
\begin{equation}
\begin{aligned}
\mathbf{q}_{0:N_{\mathrm{wp}}}^{\star}
=
\underset{\mathbf{q}_{0:N_{\mathrm{wp}}}}{\operatorname{argmin}}
\quad &
\lambda_{\mathrm{task}}
\mathcal{J}_{\mathrm{task}}(\mathbf{q}_{0:N_{\mathrm{wp}}})
\\
&+
\lambda_{\mathrm{coll}}
\underbrace{
\sum_{n=1}^{N_{\mathrm{wp}}-1}
\rho(I_{n})
}_{\mathcal{J}_{\mathrm{coll}}(\mathbf{q}_{0:N_{\mathrm{wp}}})}
\\[2mm]
\text{subject to}\quad
&
P_{\mathrm{coll}}(\mathbf{p}_m(\mathbf{q});R_m)<\tau_c,
\\
&\forall\,\mathbf{q}\in\mathcal{Q}(\mathbf{q}_{0:N_{\mathrm{wp}}}),\quad m=1,\ldots,M, \\
&\mathbf{q}_0=\mathbf{q}_s,\quad \mathbf{q}_{N_{\mathrm{wp}}}=\mathbf{q}_g.
\end{aligned}
\label{eq:trajectory_optimization}
\end{equation}
Here, $\mathcal{J}_{\mathrm{task}}$ is a task-specific trajectory cost (e.g., smoothness or path length), $\lambda_{\mathrm{task}}$ and $\lambda_{\mathrm{coll}}$ weight the task and collision costs, respectively, and $\tau_c$ is the prescribed risk threshold.
The set $\mathcal{Q}(\mathbf{q}_{0:N_{\mathrm{wp}}})$ contains the waypoints and the configurations interpolated between them with a fixed step, and the constraint is enforced conservatively through the upper bound $A^{\mathrm{ub}}$ in \eqref{eq:rendered_collision_upper}.
We consider two choices for $\rho$: the mean of all $B$ pixel opacities and the mean of the $N_{\mathrm{top}}$ largest values, where $1\leq N_{\mathrm{top}}\leq B$.
The gradient of either image cost is obtained by aggregating the pixel gradients in \eqref{eq:pixel_center_gradient}; for the latter choice, the selected pixel indices are held fixed during differentiation.
The gradient with respect to the optimization-sphere center is mapped to robot configuration space through
\begin{equation}
    \nabla_{\mathbf{q}_{n}}\mathcal{J}_{\mathrm{coll}}
    =J_{\mathbf{p}_{\mathrm{o}}}(\mathbf{q}_{n})^{\mathsf{T}}
    \nabla_{\mathbf{p}_{n}}\rho(I_{n}),
    \qquad n=1,\ldots,N_{\mathrm{wp}}-1,
    \label{eq:configuration_collision_gradient}
\end{equation}
where $J_{\mathbf{p}_{\mathrm{o}}}=\partial\mathbf{p}_{\mathrm{o}}/\partial\mathbf{q}_{n}$ is the Jacobian of the optimization-sphere center.
For a mobile robot whose configuration is the sphere center, $J_{\mathbf{p}_{\mathrm{o}}}$ is the identity.
We implement gradient descent in PyTorch on the GPU, combining these collision gradients with the task-cost gradient as specified in Algorithm~\ref{alg:Traj_Opti}.
Each candidate update is verified by the collision checker using \eqref{eq:rendered_collision_upper}.
The update is accepted only if every collision sphere satisfies the prescribed risk threshold; otherwise, the previous trajectory is retained.
Since the initial path is certified and only certified updates are accepted, every iterate satisfies the risk constraint, so the optimizer can be terminated at any iteration with a feasible trajectory.
We do not claim convergence to a local optimum of \eqref{eq:trajectory_optimization}; the optimizer serves as a local refinement of the initial path.

%% file: tex/results.tex
\section{Experimental Results}
\label{Sec: experimental results}
We evaluate the proposed trajectory planning framework and its two modules, the rendering-based collision checker and the pixel-gradient-based trajectory optimizer, in simulation and real-world environments.
All experiments are conducted on a PC equipped with an Intel Core Ultra 9 285K CPU and an NVIDIA RTX 5080 GPU.

\subsection{Baseline Methods}
We compare the proposed method with four radiance field-based methods: NeRF-Nav, CATNIPS, Splat-Nav, and Splanning.
Because these methods use different scene representations, including NeRF and 3DGS, we train each representation using the procedure described in the corresponding paper.
We also evaluate two conventional collision-map representations: OctoMap and a triangle mesh.
The OctoMap representation is constructed from the trained SVR by treating each voxel as a solid cell with a single representative density and thresholding it:
\begin{equation}
    \sigma_v = \explin\!\Big(\max_{a,b,c\in\{0,1\}} Z^{(v)}_{abc}\Big), \
    \operatorname{occupied}(v) \iff \sigma_v \ge \sigma_{\mathrm{th}},
\label{eq:octomap_voxel_occupancy}
\end{equation}
where $Z^{(v)}_{abc}$ are the eight raw corner-density values of voxel $v$ and $\sigma_{\mathrm{th}}$ is the density threshold.
We evaluated the minimum, mean, and maximum corner density as the representative density and report the best-performing variant, the maximum.
For the mesh baseline, we use the TSDF fusion and marching-cubes extraction procedure presented in SVRaster~\cite{svraster} to obtain a collision mesh from the trained representation.

\begingroup
\raggedbottom
\subsection{Collision Checker Conservativeness}
\label{subsec:detector_conservativeness}

\begin{figure}[t]
    \centering
    \includegraphics[width=0.48\textwidth]{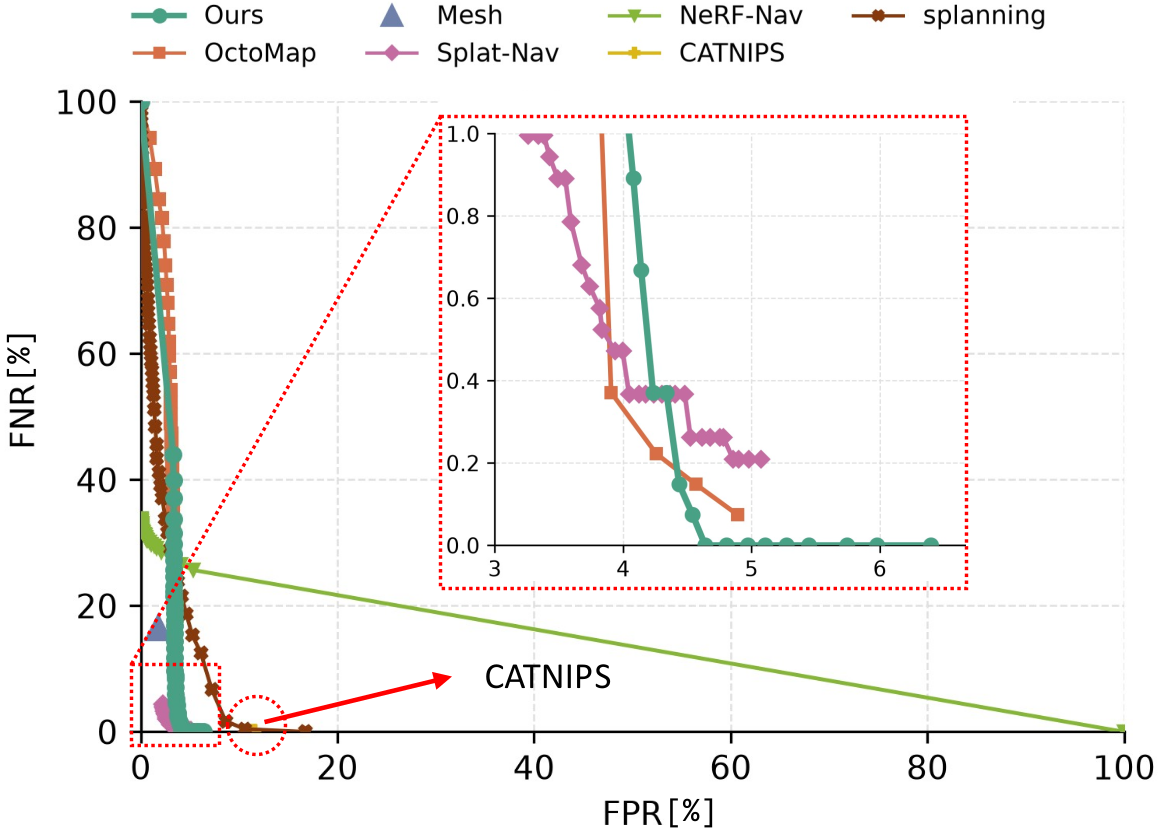}
    \caption{False-negative rate (FNR) versus false-positive rate (FPR) for collision detection. The inset magnifies the low-FNR region to show the trade-off between missed collisions and unnecessary rejection of free configurations.}
    \label{fig:FNRFPR}
\end{figure}

Equation~\eqref{eq:rendered_collision_upper} provides an upper bound on collision probability, but its error bounds may also introduce excessive conservativeness, resulting in false-positive collision detections.
We therefore evaluate the trade-off between missed collisions and unnecessary rejections using a synthetic environment containing randomly generated boxes, spheres, and cones.
Each method is used as a collision checker for a robot configuration specified by its position in $\mathbb{R}^3$.
We query the checkers on a uniform spatial grid and compare their predictions with collision labels obtained from the ground-truth (GT) mesh.
The false-negative rate (FNR) is the fraction of GT-colliding configurations classified as free, whereas the false-positive rate (FPR) is the fraction of GT-free configurations classified as colliding.
Thus, FNR measures missed collisions, while FPR measures empirical conservativeness relative to the GT geometry.

Fig.~\ref{fig:FNRFPR} shows the FNR--FPR trade-off curves obtained by varying the thresholds.
Table~\ref{tab:collision_fpr_fnr} reports the lowest observed FNR for each method and the FPR at the same operating point.
Ties in FNR are resolved by the lowest FPR.

\begin{table}[t]
\caption{Collision-checker operating points with the lowest observed FNR (\%). Ties in FNR are resolved by the lowest FPR.}
\label{tab:collision_fpr_fnr}
\centering
\tablefont
\setlength{\tabcolsep}{20pt}
\renewcommand{\arraystretch}{1.1}
\begin{tabular}{lrr}
\hline
Method & FNR (\%) & FPR (\%) \\
\hline
NeRF-Nav & 20.755 & 9.784 \\
CATNIPS & 0.000 & 11.431 \\
Splat-Nav & 0.000 & 12.254 \\
Splanning & 0.000 & 12.908 \\
OctoMap & 0.074 & 4.892 \\
Mesh & 16.493 & 1.781 \\
\rowcolor{green!15}Ours & \textbf{0.000} & \textbf{4.640} \\
\hline
\end{tabular}
\end{table}

The proposed checker attains zero observed FNR with an FPR of 4.640\%.
CATNIPS, Splat-Nav, and Splanning also attain zero observed FNR, but their corresponding FPRs are 11.431\%, 12.254\%, and 12.908\%, respectively.
At these operating points, the proposed method has the lowest FPR among checkers with no observed false negatives, reducing unnecessary rejections by 6.791--8.268 percentage points relative to these baselines.
OctoMap retains a nonzero FNR of 0.074\% with an FPR of 4.892\%.
The mesh baseline has a lower FPR of 1.781\%, but this is accompanied by an FNR of 16.493\%; NeRF-Nav has an FNR of 20.755\% at its reported operating point.
These results indicate that the upper-bound criterion preserves more free space than the other evaluated zero-FNR checkers while avoiding missed collisions on the evaluated grid.

\subsection{Mobile Robot Planning}
\label{subsec:mobile_robot_evaluation}

\begin{figure*}[t]
    \centering
    \includegraphics[width=\textwidth]{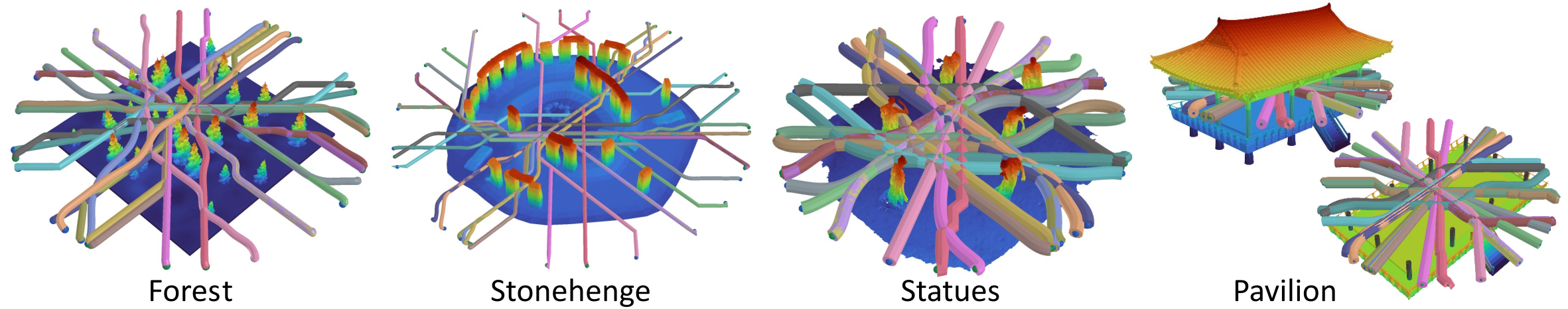}
    \caption{Trajectories generated by the proposed method in Forest, Pavilion, Statues, and Stonehenge. Each trajectory is shown in a distinct color, and its tube radius corresponds to the robot's collision-sphere radius. Map colors indicate height within each scene.}
    \label{fig:path}
\end{figure*}

We next evaluate planning performance in the Forest, Pavilion, Statues, and Stonehenge scenes.
Statues and Stonehenge follow the scene choices used in baseline studies~\cite{splatnav}.
To broaden the range of obstacle geometry, we additionally include Forest\footnote{\url{https://skfb.ly/oYLWI}} and Pavilion\footnote{\url{https://skfb.ly/oGnO6}}, constructed from publicly available mesh assets.
Forest contains fine foliage geometry, whereas Pavilion includes overhead surfaces, introducing obstacle arrangements that complement those in Statues and Stonehenge.
We model the robot as a collision sphere with radii of 0.05, 0.20, 0.10, and 0.05\,m in Forest, Pavilion, Statues, and Stonehenge, respectively.
The radii are chosen according to the scene scale and geometry to produce nontrivial planning queries.
All other hyperparameters of the proposed method are shared across scenes and listed in Table~\ref{tab:hyperparameters}.
Splanning is evaluated separately in the robot-arm experiment in Section~\ref{subsec:arm_evaluation}, since its planner targets articulated robot arms.

Each of the six mobile-robot methods is evaluated on the same 20 start--goal pairs per scene.
To assess collision-free completion, we compare the volume swept continuously by the collision sphere along every segment of each saved trajectory polyline with the GT triangle mesh.
We report collision-free completion rate, minimum obstacle distance, path length, and computation time.
The trajectories generated by the proposed method are visualized in Fig.~\ref{fig:path}.
The quantitative results below assess both collision avoidance and the conservativeness of the resulting paths.

\input{tex/tables/hyperparameters}
\input{tex/tables/success_rate}
\noindent\textbf{Collision-free completion.}
Table~\ref{tab:mobile_success_rate} shows that the proposed method passes the GT collision audit for all 80 queries and is the only method to achieve 100\% collision-free completion in every scene.
The scene-wise results vary substantially for the radiance field baselines: CATNIPS achieves 100\% in Statues but 65\% in Forest, while NeRF-Nav ranges from 20\% to 65\% across the four scenes.
Most baselines struggle in Forest.
The proposed method maintains collision-free completion across these obstacle arrangements, consistent with the use of the upper-bound criterion during search and verification of trajectory updates.

\input{tex/tables/obstacle_distance}
\input{tex/tables/path_length}
\noindent\textbf{Minimum obstacle distance and path length.}
Table~\ref{tab:mobile_obstacle_distance} reports the minimum distance from each saved trajectory's centerline to the GT obstacle surface, and Table~\ref{tab:mobile_path_length} reports path length.
The proposed method achieves the largest mean obstacle distance in Forest and Pavilion; CATNIPS achieves the largest mean distance in Statues and Stonehenge.
Meanwhile, the proposed method produces one of the shortest paths in all four scenes.
Its mean path length is the shortest in Statues and the second shortest in the other three scenes.
Together, these results support the intended balance between collision avoidance, obstacle separation, and limited path-length inflation on the evaluated queries.

\begin{figure}[t]
    \centering
    \includegraphics[width=0.48\textwidth]{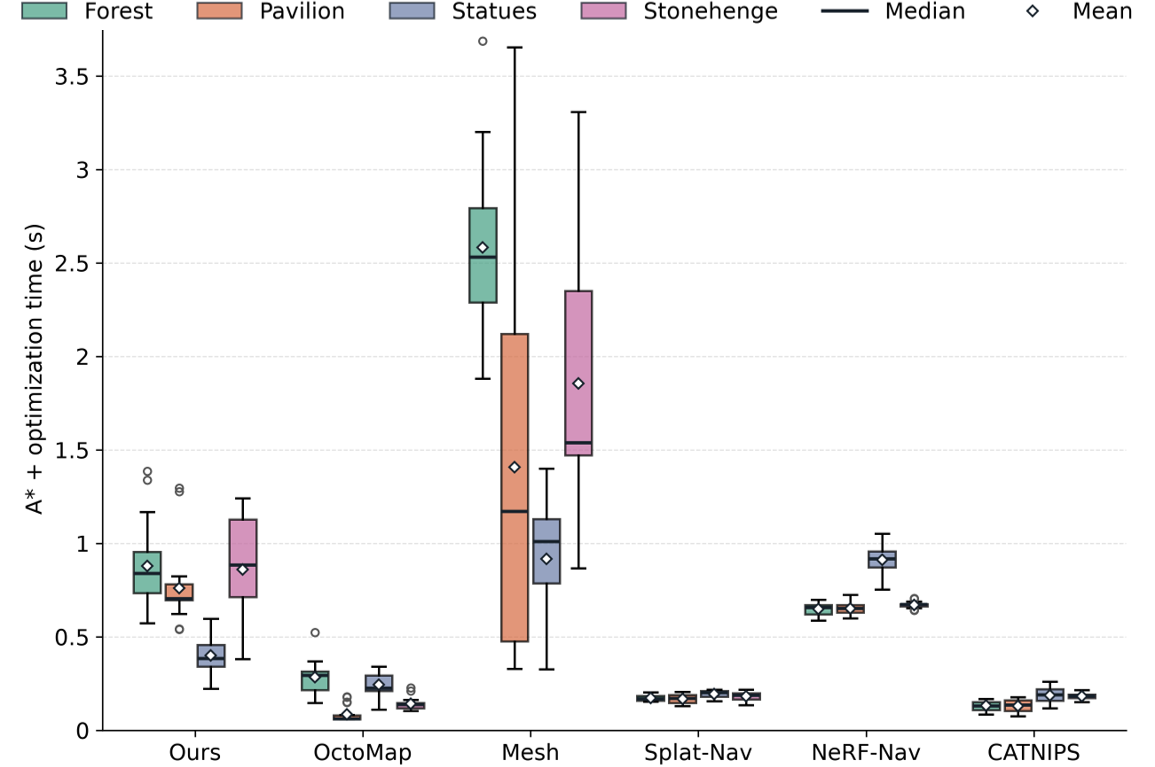}
    \caption{Combined initial-search and trajectory-optimization time for the mobile-robot planning methods, grouped by method and colored by scene. Map preprocessing times are reported separately in Table~\ref{tab:mobile_timing}.}
    \label{fig:timing_box}
\end{figure}
\input{tex/tables/timing}
\noindent\textbf{Search, optimization, and offline map preprocessing time.}
Fig.~\ref{fig:timing_box} reports the combined initial-search and optimization time, while Table~\ref{tab:mobile_timing} reports the separately logged map preprocessing time.
The proposed method requires 0.400--0.880\,s on average across the four scenes, providing subsecond mean planning times in the evaluated workloads.
The baselines require different map preprocessing procedures, including geometric collision-set and polytope construction for Splat-Nav, voxel preparation for NeRF-Nav, and robot-specific PURR construction for CATNIPS.
The separately logged map preprocessing times for these methods range from 4.542 to 22.603\,s across methods and scenes.
The proposed method operates directly on the trained sparse voxel radiance field, without constructing an auxiliary collision map.

\input{tex/tables/collision_checking_time}
\noindent\textbf{Per-call collision-checking time.}
Table~\ref{tab:mobile_collision_check_time} reports the mean and sample standard deviation of the proposed method's collision-query times, pooling all logged planning and optimizer calls across the 20 runs in each scene.
The observed means range from 0.078 to 0.387\,ms per call, demonstrating submillisecond mean collision-query latency on the evaluated GPU and workloads.
These times include the per-query preparation of the local copy of the SVR described in Section~\ref{sec:collision}, i.e., subdivision of the enclosed voxels and filling of empty regions.
Together with the stage timings, these measurements support efficient repeated collision checking while distinguishing its cost from overall planning latency.

\endgroup

\subsection{Ablation Study}
\label{subsec:ablation_study}
To evaluate the contribution of the differentiable collision cost, we compare the full optimizer with a variant that sets $\lambda_{\mathrm{coll}}=0$ in~\eqref{eq:trajectory_optimization}.
The variant retains the task cost and the accept/reject test, so every accepted step still satisfies the risk constraint.
Both variants use the same 20 start--goal pairs per scene, robot radii, initial search, optimization-sphere radius $R_{\mathrm{o}}=0.3$\,m, and step-attempt limit.
The settings in Section~\ref{subsec:mobile_robot_evaluation} balance clearance, path length, and computation time with $\lambda_{\mathrm{coll}}=1$.
To make its effect observable, this study uses $\lambda_{\mathrm{coll}}=10$.
In addition to the minimum clearance, we report the mean clearance, computed from the centerline distances to the GT surface after densifying each trajectory at 1\,cm intervals.

\input{tex/tables/collision_cost_ablation}

Table~\ref{tab:collision_cost_ablation} shows that both variants achieve 100\% GT collision-free completion in all four scenes, consistent with the certified acceptance test.
Including the collision cost increases the mean clearance by 7.2, 1.7, 24.1, and 7.5\,mm on average in Forest, Pavilion, Statues, and Stonehenge, respectively.
A Wilcoxon signed-rank test over the paired queries shows that this increase is significant in all four scenes ($p<0.01$).
The minimum clearance increases by 3.1, 12.8, 61.1, and 5.9\,mm on average in the same scenes.
These results indicate that the collision cost consistently moves trajectories away from obstacles without sacrificing collision-free completion.

\subsection{Real-World Demonstration}
\begin{figure}[t]
    \centering
    \includegraphics[width=0.48\textwidth]{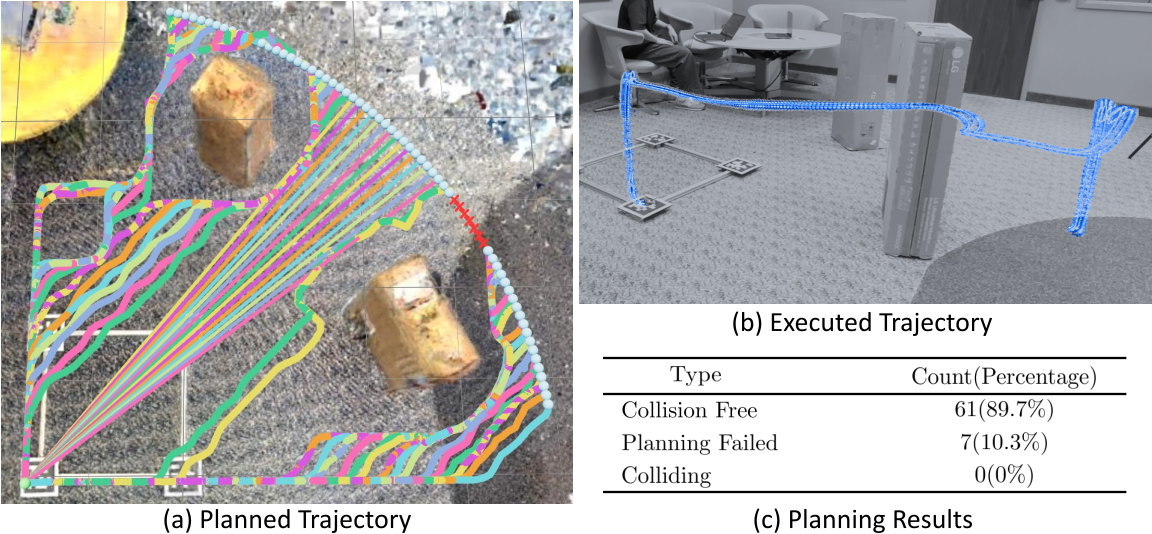}
    \caption{Real-world drone demonstration. (a) Trajectories planned by the proposed method in the SVR reconstructed from mobile-phone images, with failed goal positions marked in red. (b) Trajectory executed by a Crazyflie 2.1 drone with a motion capture system. (c) Planning outcomes over all 68 start--goal pairs.}
    \label{fig:realworld}
\end{figure}
An advantage of a radiance field map is that it can be reconstructed from camera images and then used for path planning.
To evaluate the proposed method in real-world settings, we conduct a drone experiment in an indoor environment with box obstacles.
We capture the scene with a standard mobile phone, train an SVR on the images, and plan trajectories for 68 start--goal pairs in the reconstructed scene.
Since no GT mesh is available for the real-world scene, each planned trajectory is inspected visually in the rendered SVR and checked for collision against a reference scene, built from the reconstructed mesh with the box obstacles.
As shown in Fig.~\ref{fig:realworld}(c), 61 queries (89.7\%) yield a collision-free trajectory, and no query yields a colliding one.
The remaining 7 queries (10.3\%) fail because their goal positions are classified as occupied.
These goals lie near the boundary of the captured region, where sparse image coverage leaves the reconstruction incomplete and the checker conservatively reports the region as occupied.
Fig.~\ref{fig:realworld}(a) shows the planned trajectories.
We then execute the planned trajectory on a Crazyflie 2.1 drone whose pose is estimated by an external motion capture system, as shown in Fig.~\ref{fig:realworld}(b).
The executed trajectory avoids the physical obstacles.
This experiment supports the practical feasibility of planning directly on an SVR reconstructed from camera images.

%% file: tex/tables/hyperparameters.tex
\begin{table}[t]
\centering
\tablefont
\setlength{\tabcolsep}{4pt}
\renewcommand{\arraystretch}{1.1}
\caption{Hyperparameters of the proposed method, shared by all scenes in Section~\ref{subsec:mobile_robot_evaluation}.}
\label{tab:hyperparameters}
\begin{tabular*}{\columnwidth}{@{\extracolsep{\fill}}llc@{}}
\hline
Parameter & Symbol & Value \\
\hline
\multicolumn{3}{l}{\textit{Collision checking}} \\
Rendering resolution & $W\times H$ & $512\times512$ \\
Risk threshold & $\tau_c$ & 0.01 \\
Radial bins & $N_r$ & 32 \\
Path Planner & $\mathcal{P}$ & A* \\
\hline
\multicolumn{3}{l}{\textit{Trajectory optimization}} \\
Collision-cost weight & $\lambda_{\mathrm{coll}}$ & 1 (10 in Section~\ref{subsec:ablation_study}) \\
Smoothness weight & $\lambda_{\mathrm{task}}$ & 1 \\
Image-cost aggregator & $\rho$ & top-32 mean \\
Optimization-sphere radius & $R_{\mathrm{o}}$ & 0.3\,m \\
\hline
\end{tabular*}
\end{table}

%% file: tex/tables/success_rate.tex
\begin{table}[H]
\centering
\tablefont
\setlength{\tabcolsep}{9pt}
\renewcommand{\arraystretch}{1.1}
\caption{Collision-free success rate (\%). A run is successful if planning completes and its saved trajectory has no ground-truth collision. Cell backgrounds mark the best (green), second-best (yellow), and third-best (red) values per scene.}
\label{tab:mobile_success_rate}
\begin{tabular}{lcccc}
\hline
Method & Forest & Pavilion & Statues & Stonehenge \\
\hline
Ours & \cellcolor{green!15}\textbf{100.0} & \cellcolor{green!15}\textbf{100.0} & \cellcolor{green!15}\textbf{100.0} & \cellcolor{green!15}\textbf{100.0} \\
OctoMap & \cellcolor{yellow!25}80.0 & \cellcolor{green!15}\textbf{100.0} & \cellcolor{yellow!25}95.0 & \cellcolor{yellow!25}90.0 \\
Mesh & 70.0 & \cellcolor{green!15}\textbf{100.0} & \cellcolor{green!15}\textbf{100.0} & \cellcolor{yellow!25}90.0 \\
Splat-Nav & \cellcolor{red!15}75.0 & \cellcolor{yellow!25}95.0 & \cellcolor{yellow!25}95.0 & \cellcolor{yellow!25}90.0 \\
NeRF-Nav & 20.0 & \cellcolor{red!15}65.0 & \cellcolor{red!15}25.0 & 35.0 \\
CATNIPS & 65.0 & \cellcolor{yellow!25}95.0 & \cellcolor{green!15}\textbf{100.0} & \cellcolor{red!15}70.0 \\
\hline
\end{tabular}
\end{table}

%% file: tex/tables/obstacle_distance.tex
\begin{table}[!t]
\centering
\tablefont
\setlength{\tabcolsep}{2pt}
\renewcommand{\arraystretch}{1.1}
\caption{Minimum ground-truth obstacle distance (m), reported as mean $\pm$ sample standard deviation. Cell backgrounds mark the best (green), second-best (yellow), and third-best (red) values per scene.}
\label{tab:mobile_obstacle_distance}
\begin{tabular}{lcccc}
\hline
Method & Forest & Pavilion & Statues & Stonehenge \\
\hline
Ours & \cellcolor{green!15}$\mathbf{0.07 \pm 0.01}$ & \cellcolor{green!15}$\mathbf{0.38 \pm 0.06}$ & $0.16 \pm 0.04$ & \cellcolor{yellow!25}$0.10 \pm 0.03$ \\
OctoMap & $0.05 \pm 0.01$ & \cellcolor{red!15}$0.34 \pm 0.10$ & \cellcolor{red!15}$0.16 \pm 0.05$ & $0.08 \pm 0.03$ \\
Mesh & $0.06 \pm 0.01$ & $0.33 \pm 0.10$ & \cellcolor{yellow!25}$0.17 \pm 0.05$ & \cellcolor{red!15}$0.09 \pm 0.03$ \\
Splat-Nav & \cellcolor{yellow!25}$0.06 \pm 0.02$ & $0.31 \pm 0.08$ & $0.14 \pm 0.04$ & $0.09 \pm 0.04$ \\
NeRF-Nav & $0.04 \pm 0.05$ & $0.25 \pm 0.15$ & $0.08 \pm 0.08$ & $0.05 \pm 0.04$ \\
CATNIPS & \cellcolor{red!15}$0.06 \pm 0.03$ & \cellcolor{yellow!25}$0.37 \pm 0.07$ & \cellcolor{green!15}$\mathbf{0.18 \pm 0.04}$ & \cellcolor{green!15}$\mathbf{0.10 \pm 0.09}$ \\
\hline
\end{tabular}
\end{table}

%% file: tex/tables/path_length.tex
\begin{table}[H]
\centering
\tablefont
\setlength{\tabcolsep}{2pt}
\renewcommand{\arraystretch}{1.1}
\caption{Path length (m), reported as mean $\pm$ sample standard deviation over available trajectories from 20 attempted runs per method and scene. Cell backgrounds mark the best (green), second-best (yellow), and third-best (red) values per scene.}
\label{tab:mobile_path_length}
\begin{tabular}{lcccc}
\hline
Method & Forest & Pavilion & Statues & Stonehenge \\
\hline
Ours & \cellcolor{yellow!25}$5.32 \pm 0.14$ & \cellcolor{yellow!25}$10.29 \pm 0.32$ & \cellcolor{green!15}$\mathbf{5.63 \pm 0.11}$ & \cellcolor{yellow!25}$11.00 \pm 0.48$ \\
OctoMap & \cellcolor{green!15}$\mathbf{5.31 \pm 0.14}$ & \cellcolor{green!15}$\mathbf{10.23 \pm 0.30}$ & \cellcolor{yellow!25}$5.65 \pm 0.17$ & \cellcolor{green!15}$\mathbf{10.57 \pm 0.30}$ \\
Mesh & $5.76 \pm 0.36$ & $10.94 \pm 1.10$ & $6.09 \pm 0.49$ & $11.58 \pm 0.92$ \\
Splat-Nav & $5.52 \pm 0.32$ & $11.55 \pm 1.02$ & $5.99 \pm 0.44$ & $11.42 \pm 0.86$ \\
NeRF-Nav & $5.83 \pm 0.57$ & $11.87 \pm 1.08$ & $6.23 \pm 0.55$ & $12.00 \pm 0.71$ \\
CATNIPS & \cellcolor{red!15}$5.47 \pm 0.36$ & \cellcolor{red!15}$10.90 \pm 0.82$ & \cellcolor{red!15}$5.75 \pm 0.37$ & \cellcolor{red!15}$11.39 \pm 0.65$ \\
\hline
\end{tabular}
\end{table}

%% file: tex/tables/timing.tex
\begin{table}[H]
\centering
\tablefont
\setlength{\tabcolsep}{1pt}
\renewcommand{\arraystretch}{1.1}
\caption{Map preprocessing time (s), reported as mean $\pm$ sample standard deviation over 20 runs per scene.}
\label{tab:mobile_timing}
\begin{tabular}{lcccc}
\hline
Method & Forest & Pavilion & Statues & Stonehenge \\
\hline
Ours & \cellcolor{green!15}\textbf{0} & \cellcolor{green!15}\textbf{0} & \cellcolor{green!15}\textbf{0} & \cellcolor{green!15}\textbf{0} \\
Splat-Nav & $5.312 \pm 0.084$ & $5.262 \pm 0.082$ & $7.989 \pm 0.083$ & $5.425 \pm 0.089$ \\
NeRF-Nav & $4.667 \pm 0.052$ & $4.542 \pm 0.058$ & $5.592 \pm 0.250$ & $4.713 \pm 0.043$ \\
CATNIPS & $5.026 \pm 0.055$ & $6.920 \pm 0.050$ & $22.603 \pm 0.444$ & $5.449 \pm 0.075$ \\
\hline
\end{tabular}
\end{table}

%% file: tex/tables/collision_checking_time.tex
\begin{table}[H]
\centering
\tablefont
\setlength{\tabcolsep}{2pt}
\renewcommand{\arraystretch}{1.1}
\caption{Collision-checking time [ms], reported as mean $\pm$ sample standard deviation over all logged calls from 20 runs per scene.}
\label{tab:mobile_collision_check_time}
\begin{tabular}{lcccc}
\hline
Method & Forest & Pavilion & Statues & Stonehenge \\
\hline
Ours & $0.387 \pm 2.671$ & $0.078 \pm 1.768$ & $0.168 \pm 2.112$ & $0.129 \pm 1.662$ \\
\hline
\end{tabular}
\end{table}

%% file: tex/tables/collision_cost_ablation.tex
\begin{table}[!t]
\centering
\tablefont
\setlength{\tabcolsep}{3pt}
\renewcommand{\arraystretch}{1.1}
\caption{Collision-cost ablation over 20 start--goal pairs per scene. Both variants use identical search and optimizer settings and differ only in whether $\lambda_{\mathrm{coll}}$ is zero. Clearance is the distance from the trajectory centerline to the GT surface. Continuous metrics are mean $\pm$ sample standard deviation.}
\label{tab:collision_cost_ablation}
\begin{tabular}{llccc}
\hline
Scene & $\mathcal{J}_{\mathrm{coll}}$ & Free (\%) & Min. clear. (m) & Mean clear. (m) \\
\hline
Forest & off & 100 & $0.071 \pm 0.009$ & $0.322 \pm 0.054$ \\
 & on & 100 & $\mathbf{0.074 \pm 0.020}$ & $\mathbf{0.329 \pm 0.053}$ \\
\hline
Pavilion & off & 100 & $0.364 \pm 0.067$ & $0.750 \pm 0.145$ \\
 & on & 100 & $\mathbf{0.376 \pm 0.054}$ & $\mathbf{0.751 \pm 0.144}$ \\
\hline
Statues & off & 100 & $0.163 \pm 0.053$ & $0.468 \pm 0.031$ \\
 & on & 100 & $\mathbf{0.224 \pm 0.079}$ & $\mathbf{0.492 \pm 0.025}$ \\
\hline
Stonehenge & off & 100 & $0.126 \pm 0.045$ & $0.603 \pm 0.047$ \\
 & on & 100 & $\mathbf{0.132 \pm 0.045}$ & $\mathbf{0.611 \pm 0.047}$ \\
\hline
\end{tabular}
\end{table}

%% file: tex/discussion.tex
\section{Discussion}
This section discusses the applicability and limitations of the proposed method. We first apply the collision checker to an articulated robot arm, and then clarify what the collision bound does and does not guarantee.

\subsection{Extension to Articulated Robot Arms}
\label{subsec:arm_evaluation}
Although the main target of the proposed method is mobile robots such as drones, the collision checker does not depend on a particular planner or a single rigid sphere.
To examine this extensibility, we apply it to a 6-DOF UR10e arm without arm-specific design or tuning.
The arm is approximated by 17 collision spheres attached to its links, and each configuration is checked by rendering every sphere with the proposed criterion.
The checker is plugged into the RRT-Connect~\cite{kuffner2000rrtconnect} planner in OMPL~\cite{sucan2012ompl}, validating each edge at 0.05\,rad, followed by the native path simplification of OMPL, whose shortcuts are also validated by the checker.
Trajectory optimization is not applied.
As a reference, we run Splanning~\cite{splanning}, which plans on a 3DGS model of the same scene with a receding-horizon trajectory optimizer and uses its own collision model of the arm.

\begin{figure}[t]
    \centering
    \includegraphics[width=0.48\textwidth]{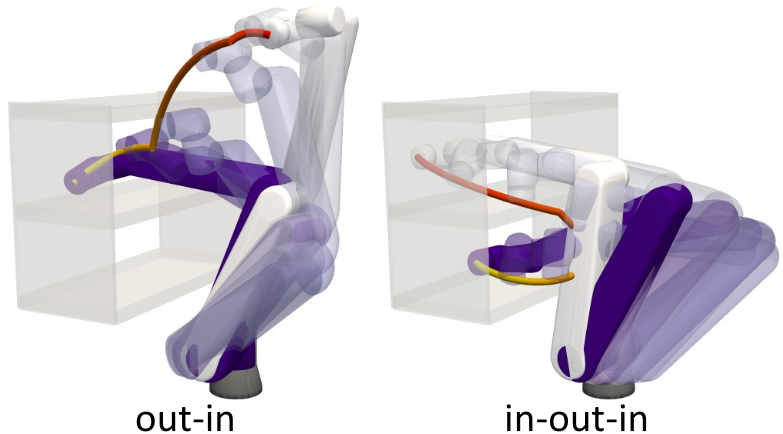}
    \caption{Representative trajectories planned by the proposed method for the out--in (left) and in--out--in (right) tasks. Robot poses are sampled evenly along each path and colored from start (dark) to goal (light), and the tube shows the end-effector path.}
    \label{fig:arm_path}
\end{figure}

We consider two tasks in a shelf scene.
In the out--in task, the arm moves from a configuration outside the shelf to one inside a compartment.
In the in--out--in task, the arm moves between two compartments, which requires leaving the first compartment before entering the second.
For each task, we run 50 trials in which every joint of the nominal start and goal configurations is perturbed by Gaussian noise with a standard deviation of 0.1\,rad.
Each resulting trajectory is checked for collision against the GT mesh.
Fig.~\ref{fig:arm_path} shows representative trajectories of the proposed method for both tasks.

\input{tex/tables/arm_results}

As shown in Table~\ref{tab:arm_results}, the proposed checker yields GT collision-free trajectories in all planned trials in this setup, with success rates of 100\% and 98\%.
The remaining in--out--in trial reached the timeout.
Splanning succeeds in 64\% and 42\% of the trials, respectively.
Most of its failures stop short of the goal, because its objective reduces the distance to the goal at each step and therefore becomes trapped in local minima when the arm must first move away from the goal, as in the in--out--in task.
In addition, 1 and 6 of its trajectories collide with the GT mesh in the two tasks.
Splanning requires less planning time on the trials it solves, whereas the proposed method spends most of its time on collision checks within the sampling-based search.

These observations suggest that the collision checker can be used for articulated robots by combining it with an off-the-shelf planner.
In this setting, however, the planning time grows linearly with the number of spheres.
A dedicated study, including joint-space trajectory optimization through the sphere kinematics and a broader set of scenes and tasks, remains future work.

\subsection{Scope of the Collision Bound}
The proposed criterion bounds the collision probability of every radial segment of each collision sphere individually, rather than the joint probability that at least one segment collides.
A joint probability cannot be obtained by treating directions as independent, since a single obstacle occludes a bundle of neighboring rays, and assuming independence over a continuum of directions or over rendered pixels makes the result degenerate or resolution dependent.
A joint formulation instead requires modeling the particle scale, since the density of a radiance field corresponds to the product of particle number density and cross-sectional area~\cite{max1995optical}; this scale is not identifiable from a radiance field trained on images, and the proposed per-direction measure avoids it.

The bound applies only to the configurations at which the collision checker is evaluated.
Following common practice in sampling-based planning, motion between these configurations is validated at a finite resolution rather than certified~\cite{elbanhawi2014sampling}.
Certifying continuous motion by radius inflation does not follow directly, because the radial segments of a sphere at an intermediate position are not radial segments of the inflated sphere, so the per-direction measure is not monotone under sphere inclusion.
As in other radiance field-based planners, the bound is stated with respect to the reconstructed scene.

Empirically, the per-direction criterion proved sufficient: it yielded no missed collisions at the reported operating point on the evaluated query grid (Table~\ref{tab:collision_fpr_fnr}), produced ground-truth collision-free trajectories in all 80 mobile-robot planning queries (Table~\ref{tab:mobile_success_rate}), and avoided all obstacles in the real-world experiment.
Extending the bound to the joint collision probability of a sphere, of the full robot, or of an entire trajectory, and to continuous motion, remains future work.

%% file: tex/tables/arm_results.tex
\begin{table}[H]
\centering
\tablefont
\setlength{\tabcolsep}{2pt}
\renewcommand{\arraystretch}{1.1}
\caption{Robot-arm planning results over 50 trials per task. Success requires reaching the goal with a GT collision-free trajectory. Not reached counts trials that stopped in a local minimum, reached the step limit, or timed out without a GT collision. Planning time is mean $\pm$ sample standard deviation over successful trials.}
\label{tab:arm_results}
\begin{tabular}{llcccc}
\hline
Task & Method & \begin{tabular}[c]{@{}c@{}}Success\\(\%)\end{tabular} & \begin{tabular}[c]{@{}c@{}}GT coll.\\(\%)\end{tabular} & \begin{tabular}[c]{@{}c@{}}Not reached\\(\%)\end{tabular} & \begin{tabular}[c]{@{}c@{}}Time\\(s)\end{tabular} \\
\hline
Out--in & Ours & \textbf{100} & \textbf{0} & \textbf{0} & $6.98 \pm 5.57$ \\
 & Splanning & 64 & 2 & 34 & $\mathbf{2.35 \pm 0.87}$ \\
\hline
In--out--in & Ours & \textbf{98} & \textbf{0} & \textbf{2} & $20.72 \pm 18.52$ \\
 & Splanning & 42 & 12 & 46 & $\mathbf{2.44 \pm 0.67}$ \\
\hline
\end{tabular}
\end{table}

%% file: tex/conclusion.tex
\section{Conclusion}
This paper presented Collision Rendering, a trajectory planning framework that operates directly on a trained sparse voxel radiance field.
Its central observation is that the rendering pipeline of SVR can serve as a numerical approximator of collision probability.
Under a transmittance-based collision model, we derived an upper bound on the maximum directional collision probability of a collision sphere from a spherical opacity image and error bounds computed from voxel-wise Lipschitz bounds.
Evaluating this bound on the GPU yields a modular collision checker that can be used by any configuration-based planner.
The same rasterization also provides a differentiable collision cost, whose analytic pixel-opacity gradient refines planned trajectories.

In simulation, the checker attained zero observed false negatives with the lowest false-positive rate among the zero-FNR checkers, at millisecond mean query times.
The resulting planner produced GT collision-free trajectories in all 80 mobile-robot queries with subsecond mean planning times, and the ablation showed that the collision cost increases clearance.
Without arm-specific design, the checker also enabled an off-the-shelf sampling-based planner to solve robot-arm tasks.
In a real-world drone experiment, planning on an SVR reconstructed from mobile-phone images yielded no colliding trajectory, and the executed trajectory avoided all physical obstacles.

The bound currently certifies each radial segment of a collision sphere at discrete configurations.
Future work includes extending it to the joint collision probability of a sphere, a robot, or an entire trajectory and to continuous motion, as well as joint-space trajectory optimization for articulated robots and reducing the cost of checking many collision spheres.

%% file: tex/appendix_proofs.tex
\section{Collision-Bound Derivations and Proofs}
\label{app:collision_proofs}

\subsection{Per-Ray Upper Bound}
\label{subsec:per_ray_bound}
We derive a per-ray upper bound for the integrated density approximated by rendering in \eqref{eq:pixel_segment_opacity}.

Consider a sampled pixel direction $\mathbf{d}$.
Let $\mathcal{S}(\mathbf{d})$ denote the ray--voxel intersection segments within radius $R$, as defined in Section~\ref{subsec:rendering_approximation}; zero-length segments are omitted.
For segment $s\in\mathcal{S}(\mathbf{d})$, let $[t_s^{\mathrm{in}},t_s^{\mathrm{out}}]\subseteq[0,R]$ be its radial interval.
Its entry and exit points are $\mathbf{x}_{s}^{\mathrm{in}}=\mathbf{r}_{\mathbf{p},\mathbf{d}}(t_s^{\mathrm{in}})$ and $\mathbf{x}_{s}^{\mathrm{out}}=\mathbf{r}_{\mathbf{p},\mathbf{d}}(t_s^{\mathrm{out}})$, respectively. Its physical length is
\begin{equation}
    \ell_s
    =
    \left\|
    \mathbf{x}_{s}^{\mathrm{out}}
    -
    \mathbf{x}_{s}^{\mathrm{in}}
    \right\|_{2}
    =t_s^{\mathrm{out}}-t_s^{\mathrm{in}}.
    \label{eq:segment_length}
\end{equation}
We parameterize the segment by the physical distance $t$ from the sphere center along the ray as
\begin{equation}
    \mathbf{x}_s(t)
    =
    \mathbf{r}_{\mathbf{p},\mathbf{d}}(t)
    =
    \mathbf{p}+t\mathbf{d},
    \qquad
    t\in[t_s^{\mathrm{in}},t_s^{\mathrm{out}}].
    \label{eq:segment_parameterization}
\end{equation}

Because $Z$ is trilinear inside the voxel and
$\mathbf{x}_s(t)$ is affine in $t$, the raw density restricted to the segment is a cubic polynomial:
\begin{equation}
    p_s(t)
    =
    Z(\mathbf{x}_s(t))
    =
    c_{s,0}
    +
    c_{s,1}t
    +
    c_{s,2}t^{2}
    +
    c_{s,3}t^{3}.
    \label{eq:ray_cubic}
\end{equation}
Its derivative is
\begin{equation}
    p_s'(t)
    =
    c_{s,1}
    +
    2c_{s,2}t
    +
    3c_{s,3}t^{2}.
    \label{eq:ray_cubic_derivative}
\end{equation}

\begin{lemma}[Ray-segment Lipschitz constant]
\label{lem:tight_ray_lipschitz}
The Lipschitz constant of the raw density field $Z$ along segment $s$, with respect to physical distance, is
\begin{equation}
    L_{\mathrm{ray},s}
    =
    \max_{t\in\mathcal{C}_s}
    \left|
    c_{s,1}
    +
    2c_{s,2}t
    +
    3c_{s,3}t^{2}
    \right|,
    \label{eq:tight_ray_lipschitz}
\end{equation}
where
\begin{equation}
    \mathcal{C}_s
    =
    \{t_s^{\mathrm{in}},t_s^{\mathrm{out}}\}
    \cup
    \left\{
    -\frac{c_{s,2}}{3c_{s,3}}
    \;\middle|\;
    c_{s,3}\neq 0,\,
    -\frac{c_{s,2}}{3c_{s,3}}\in[t_s^{\mathrm{in}},t_s^{\mathrm{out}}]
    \right\}.
    \label{eq:ray_lipschitz_candidates}
\end{equation}
Moreover, $L_{\mathrm{ray},s}$ is also a Lipschitz constant for the density $\sigma$ along the same segment.
\end{lemma}

\begin{proof}
Since $t$ is the physical distance along the ray,
\[
    \frac{\mathrm{d}}{\mathrm{d}t}Z(\mathbf{r}_{\mathbf{p},\mathbf{d}}(t))
    =
    p_s'(t),
\]
and therefore
\[
    L_{\mathrm{ray},s}
    =
    \max_{t\in[t_s^{\mathrm{in}},t_s^{\mathrm{out}}]}
    |p_s'(t)|.
\]
Since $p_s'(t)$ is quadratic, the maximum of its absolute value over $[t_s^{\mathrm{in}},t_s^{\mathrm{out}}]$ occurs at an endpoint or at the quadratic vertex
\[
    t^{\star}
    =
    -\frac{c_{s,2}}{3c_{s,3}},
\]
when the vertex lies inside $[t_s^{\mathrm{in}},t_s^{\mathrm{out}}]$. This gives
\eqref{eq:tight_ray_lipschitz}--\eqref{eq:ray_lipschitz_candidates}.

Finally, by \eqref{eq:explin_gradient_bound},
\[
\begin{aligned}
    \left|
    \frac{\mathrm{d}}{\mathrm{d}t}
    \sigma(\mathbf{r}_{\mathbf{p},\mathbf{d}}(t))
    \right|
    &=
    \left|
    \operatorname{explin}'(p_s(t))p_s'(t)
    \right| \\
    &\le L_{\mathrm{ray},s}.
\end{aligned}
\]
Thus, $L_{\mathrm{ray},s}$ is also a valid density Lipschitz constant.
\end{proof}

During rendering, density is sampled at $K_s\ge 1$ points along each segment.
Let $h_s=\ell_s/K_s$ be the physical sample spacing, and let
\[
    \Delta_{s,k}=[t_s^{\mathrm{in}}+(k-1)h_s,\;t_s^{\mathrm{in}}+kh_s],
    \qquad k=1,\ldots,K_s,
\]
be the corresponding subintervals.
For a sample distance $t_{s,k}\in \Delta_{s,k}$, its world-space position is $\mathbf{x}_{s,k}=\mathbf{r}_{\mathbf{p},\mathbf{d}}(t_{s,k})$.
The midpoint samples used in Section~\ref{sec:robot_center_alpha_gradient} satisfy $t_{s,k}=t_s^{\mathrm{in}}+(k-1/2)h_s$.
The sampled integrated density in \eqref{eq:pixel_segment_opacity} is therefore
\begin{equation}
    \widehat{\tau}_s
    =
    h_s
    \sum_{k=1}^{K_s}
    \sigma(\mathbf{x}_{s,k}).
    \label{eq:sampled_segment_integral}
\end{equation}

\begin{theorem}[Per-ray upper bound]
\label{thm:per_ray_upper_bound}
For sampled direction $\mathbf{d}_b$, the exact integrated density $\tau(\mathbf{d}_b)$ satisfies
\begin{equation}
    \tau(\mathbf{d}_b)
    \le
    U_{\mathrm{ray},b},
    \label{eq:per_ray_upper_statement}
\end{equation}
where
\begin{equation}
    U_{\mathrm{ray},b}
    =
    \sum_{s\in\mathcal{S}(\mathbf{d}_b)}
    \left[
        \widehat{\tau}_s
        +
        \frac{
        L_{\mathrm{ray},s}\ell_s^{2}
        }{
        2K_s
        }
    \right].
    \label{eq:per_ray_upper_bound}
\end{equation}
\end{theorem}

\begin{proof}
Set $\mathbf{d}=\mathbf{d}_b$. For a subinterval $\Delta_{s,k}$ of length $h_s$ and its sample distance $t_{s,k}$, the Lipschitz property gives
\[
    \left|
    \sigma(\mathbf{r}_{\mathbf{p},\mathbf{d}}(t))
    -
    \sigma(\mathbf{x}_{s,k})
    \right|
    \le
    L_{\mathrm{ray},s}
    |t-t_{s,k}|.
\]
Thus,
\begin{align}
    &
    \left|
    \int_{\Delta_{s,k}}
    \sigma(\mathbf{r}_{\mathbf{p},\mathbf{d}}(t))\,\mathrm{d}t
    -
    h_s
    \sigma(\mathbf{x}_{s,k})
    \right|
    \nonumber\\
    &\qquad
    \le
    L_{\mathrm{ray},s}
    \int_{\Delta_{s,k}}
    |t-t_{s,k}|\,\mathrm{d}t
    \le
    \frac{
    L_{\mathrm{ray},s}h_s^{2}
    }{2}.
    \label{eq:subinterval_quadrature_error}
\end{align}
Summing over all $K_s$ subintervals yields
\[
    \int_{t_s^{\mathrm{in}}}^{t_s^{\mathrm{out}}}\sigma(\mathbf{r}_{\mathbf{p},\mathbf{d}}(t))\,\mathrm{d}t
    \le
    \widehat{\tau}_s
    +
    \frac{
    L_{\mathrm{ray},s}\ell_s^{2}
    }{
    2K_s
    }.
\]
Summing over all ray--voxel segments proves
\eqref{eq:per_ray_upper_bound}.
\end{proof}

\subsection{Angular Upper Bound}
\label{subsec:angular_bound}
The per-ray bound in Theorem~\ref{thm:per_ray_upper_bound} holds only for the sampled pixel-center directions $\mathbf{d}_b$.
To bound $P_{\mathrm{coll}}(\mathbf{p};R)$ in \eqref{eq:coll_prob_def}, we must also bound the rays between them.

To bound rays that are not explicitly sampled, we divide the radial interval into $N_r$ radial shells
\[
    0=r_0<r_1<\cdots<r_{N_r}=R.
\]
For angular bin $\mathcal{B}_b$ in \eqref{eq:angular_bin_radius} and radial shell $[r_j,r_{j+1}]$, we define the chord-swept region
\begin{equation}
\begin{aligned}
    \widetilde{\mathcal{F}}_{b,j}
    =
    \bigg\{
    &(1-\lambda)
    (\mathbf{p}+r\mathbf{d}_b)
    +
    \lambda
    (\mathbf{p}+r\mathbf{d})
    :
    \\
    &\mathbf{d}\in \mathcal{B}_b,\;
    r\in[r_j,r_{j+1}],\;
    \lambda\in[0,1]
    \bigg\}.
\end{aligned}
\label{eq:chord_swept_region}
\end{equation}

Let
\begin{equation}
    \mathcal{V}_{b,j}
    =
    \left\{
    v
    :
    v\cap
    \widetilde{\mathcal{F}}_{b,j}
    \neq\emptyset
    \right\}
    \label{eq:angular_voxel_set}
\end{equation}
be the set of all voxels intersecting this region, and define
\begin{equation}
    G_{b,j}
    =
    \max_{v\in\mathcal{V}_{b,j}}
    L_{\mathrm{vox},v}.
    \label{eq:local_angular_gradient}
\end{equation}
\begin{lemma}[Local angular deviation bound]
\label{lem:angular_deviation}
For every direction $\mathbf{d}\in \mathcal{B}_b$,
\begin{equation}
    \left|
    \tau(\mathbf{d})
    -
    \tau(\mathbf{d}_b)
    \right|
    \le
    \varepsilon_{\mathrm{ang},b},
    \label{eq:angular_deviation_statement}
\end{equation}
where
\begin{equation}
    \varepsilon_{\mathrm{ang},b}
    =
    2\sin\left(\frac{\delta_b}{2}\right)
    \sum_{j=0}^{N_r-1}
    G_{b,j}
    \frac{
    r_{j+1}^{2}-r_j^{2}
    }{2}.
    \label{eq:local_angular_error}
\end{equation}
\end{lemma}

\begin{proof}
For any $\mathbf{d}\in \mathcal{B}_b$,
\begin{align}
    \left|
    \tau(\mathbf{d})
    -
    \tau(\mathbf{d}_b)
    \right|
    &\le
    \int_{0}^{R}
    \left|
    \sigma(\mathbf{p}+r\mathbf{d})
    -
    \sigma(\mathbf{p}+r\mathbf{d}_b)
    \right|
    \mathrm{d}r.
    \label{eq:angular_difference_integral}
\end{align}
For $r\in[r_j,r_{j+1}]$, the line segment connecting
$\mathbf{p}+r\mathbf{d}$ and
$\mathbf{p}+r\mathbf{d}_b$ lies in
$\widetilde{\mathcal{F}}_{b,j}$.
Therefore, applying the mean-value inequality within each voxel and using continuity across voxel interfaces, together with \eqref{eq:local_angular_gradient}, yields
\begin{align}
    &
    \left|
    \sigma(\mathbf{p}+r\mathbf{d})
    -
    \sigma(\mathbf{p}+r\mathbf{d}_b)
    \right|
    \nonumber\\
    &\qquad
    \le
    G_{b,j}
    \left\|
    r\mathbf{d}
    -
    r\mathbf{d}_b
    \right\|_2
    =
    G_{b,j}r
    \|\mathbf{d}-\mathbf{d}_b\|_2.
    \label{eq:angular_mean_value}
\end{align}
For two unit vectors,
\[
    \|\mathbf{d}-\mathbf{d}_b\|_2
    =
    2\sin\left(
    \frac{
    d_{\mathbb{S}^{2}}(\mathbf{d},\mathbf{d}_b)
    }{2}
    \right).
\]
Using \eqref{eq:angular_bin_radius},
\[
    \|\mathbf{d}-\mathbf{d}_b\|_2
    \le
    2\sin\left(\frac{\delta_b}{2}\right).
\]
Hence,
\[
    \left|
    \sigma(\mathbf{p}+r\mathbf{d})
    -
    \sigma(\mathbf{p}+r\mathbf{d}_b)
    \right|
    \le
    2\sin\left(\frac{\delta_b}{2}\right)
    G_{b,j}r.
\]
Integrating this inequality over each radial shell and summing gives
\begin{align}
    \left|
    \tau(\mathbf{d})
    -
    \tau(\mathbf{d}_b)
    \right|
    &\le
    2\sin\left(\frac{\delta_b}{2}\right)
    \sum_{j=0}^{N_r-1}
    G_{b,j}
    \int_{r_j}^{r_{j+1}}r\,\mathrm{d}r
    \nonumber\\
    &=
    2\sin\left(\frac{\delta_b}{2}\right)
    \sum_{j=0}^{N_r-1}
    G_{b,j}
    \frac{
    r_{j+1}^{2}-r_j^{2}
    }{2}.
\end{align}
This proves \eqref{eq:local_angular_error}.
\end{proof}

\subsection{Proof of the Sphere-Wise Upper Bound}
\label{app:proof_sphere_bound}
\begin{proof}[Proof of Theorem~\ref{thm:sphere_max_upper_bound}]
Let
\[
    \mathbf{d}^{\star}
    \in
    \arg\max_{\mathbf{d}\in\mathbb{S}^{2}}
    \tau(\mathbf{d}).
\]
Because the angular bins cover $\mathbb{S}^{2}$, there exists an index $b^{\star}$ such that
\[
    \mathbf{d}^{\star}\in\mathcal{B}_{b^{\star}}.
\]
By Lemma~\ref{lem:angular_deviation},
\[
    \tau(\mathbf{d}^{\star})
    \le
    \tau(\mathbf{d}_{b^{\star}})
    +
    \varepsilon_{\mathrm{ang},b^{\star}}.
\]
By Theorem~\ref{thm:per_ray_upper_bound},
\[
    \tau(\mathbf{d}_{b^{\star}})
    \le
    U_{\mathrm{ray},b^{\star}}.
\]
Therefore,
\[
    \tau_{\max}
    =
    \tau(\mathbf{d}^{\star})
    \le
    U_{\mathrm{ray},b^{\star}}
    +
    \varepsilon_{\mathrm{ang},b^{\star}}
    =
    U_{b^{\star}}
    \le
    \max_b U_b.
\]
Finally, $1-\exp(-x)$ is monotonically increasing, so
\[
    P_{\mathrm{coll}}(\mathbf{p};R)
    =
    1-\exp(-\tau_{\max})
    \le
    1-\exp\left(-\max_b U_b\right).
\]
\end{proof}

\subsection{Implementation Notes}
\input{tex/pseudocode_ray_lipschitz}

\input{tex/pseudocode_angular_error}

\section{Trajectory-Optimization Derivations}
\label{app:optimization_proofs}

\subsection{Gradient Setup}
To optimize a robot trajectory using rendered opacity as a collision cost, we derive the gradient of the pixel opacity $A_b$ with respect to the sphere center $\mathbf{p}$.
Let the sphere be centered at $\mathbf{p}\in\mathbb{R}^3$ with rendering radius $R$; we consider translation of $\mathbf{p}$ while holding the spherical-camera orientation fixed.
For the optimization cost, $R$ is set to the radius $R_{\mathrm{o}}$ of the optimization sphere used in Section~\ref{sec:robot_center_alpha_gradient} and \eqref{eq:trajectory_optimization}.
For spherical-image pixel $b$, let $\mathbf{d}_b$ denote its world-space unit ray direction.
The ray is parameterized as
\begin{equation}
    \mathbf{r}_{\mathbf{p},\mathbf{d}_b}(t)
    =\mathbf{p}+t\mathbf{d}_b,
    \qquad
    \frac{\partial\mathbf{d}_b}{\partial\mathbf{p}}=\mathbf{0}.
    \label{eq:spherical_ray_robot_center}
\end{equation}

The rendered pixel opacity along this ray is $A_b\in[0,1]$, and we derive its gradient $\nabla_{\mathbf{p}}A_b$, which is used as the spatial gradient of the collision cost.

\subsection{Derivation and Proof of the Voxel Gradient}
\label{app:proof_voxel_gradient}
\subsubsection{Ray--Voxel Interval}
We first restate the ray--voxel intersection segment in parameterized form.
Consider segment $s\in\mathcal{S}(\mathbf{d}_b)$ in voxel $v$, whose minimum and maximum corners are $\mathbf{m}_v$ and $\mathbf{M}_v$.
For a nonparallel coordinate axis $i\in\{x,y,z\}$, with $d_{b,i}\neq0$, the ray intersects the two voxel planes at
\begin{equation}
    t^{(0)}_{v,i}=\frac{m_{v,i}-p_i}{d_{b,i}},
    \qquad
    t^{(1)}_{v,i}=\frac{M_{v,i}-p_i}{d_{b,i}}.
    \label{eq:axis_plane_intersections}
\end{equation}
The near and far intersections along each nonparallel axis are
\begin{equation}
    n_i=\min\left(t^{(0)}_{v,i},t^{(1)}_{v,i}\right),
    \qquad
    f_i=\max\left(t^{(0)}_{v,i},t^{(1)}_{v,i}\right).
\end{equation}
The entry and exit distances for segment $s$ are
\begin{equation}
    t_s^{\mathrm{in}}=\max_i n_i,
    \qquad
    t_s^{\mathrm{out}}=\min_i f_i,
    \label{eq:ray_voxel_interval}
\end{equation}
where the extrema are taken over nonparallel axes.
These distances are then restricted to $[0,R]$, i.e., $t_s^{\mathrm{in}}\gets\max(t_s^{\mathrm{in}},0)$ and $t_s^{\mathrm{out}}\gets\min(t_s^{\mathrm{out}},R)$.
A parallel axis permits an intersection only when the ray origin lies within its voxel slab.
Let $i_{\mathrm{in}}$ and $i_{\mathrm{out}}$ denote the active axes determining the entry and exit distances, respectively.
For endpoints determined by voxel planes, holding these active axes fixed gives
\begin{equation}
\begin{aligned}
    \gamma_s^{\mathrm{in}}:=\nabla_{\mathbf{p}}t_s^{\mathrm{in}}
    &=-\frac{1}{d_{b,i_{\mathrm{in}}}}\mathbf{e}_{i_{\mathrm{in}}}, \\
    \gamma_s^{\mathrm{out}}:=\nabla_{\mathbf{p}}t_s^{\mathrm{out}}
    &=-\frac{1}{d_{b,i_{\mathrm{out}}}}\mathbf{e}_{i_{\mathrm{out}}},
\end{aligned}
    \label{eq:interval_endpoint_gradient}
\end{equation}
where $\mathbf{e}_i$ is the Cartesian standard basis vector for axis $i$.

Only nonempty segments with $\ell_s>0$ are retained.

\subsubsection{Derivative of the Sampled Point}
For the $K_s$ midpoint samples in segment $s$, indexed by $k=1,\ldots,K_s$, define the physical sample spacing and normalized midpoint positions as
\begin{equation}
    h_s=\frac{\ell_s}{K_s},
    \qquad
    u_{s,k}=\frac{k-1/2}{K_s}.
\end{equation}
The sample distance is
\begin{equation}
    t_{s,k}=(1-u_{s,k})t_s^{\mathrm{in}}+u_{s,k}t_s^{\mathrm{out}}.
    \label{eq:sample_distance}
\end{equation}
Holding $K_s$ fixed during differentiation gives
\begin{equation}
\begin{aligned}
    \nabla_{\mathbf{p}}h_s
    &=\frac{\gamma_s^{\mathrm{out}}-\gamma_s^{\mathrm{in}}}{K_s}, \\
    \nabla_{\mathbf{p}}t_{s,k}
    &=(1-u_{s,k})\gamma_s^{\mathrm{in}}+u_{s,k}\gamma_s^{\mathrm{out}}.
\end{aligned}
\end{equation}
The world-space sample position is
\begin{equation}
    \mathbf{x}_{s,k}=\mathbf{p}+t_{s,k}\mathbf{d}_b.
\end{equation}
Since $\partial\mathbf{d}_b/\partial\mathbf{p}=\mathbf{0}$,
\begin{equation}
    \frac{\partial\mathbf{x}_{s,k}}{\partial\mathbf{p}}
    =\mathbf{I}_3+\mathbf{d}_b(\nabla_{\mathbf{p}}t_{s,k})^{\mathsf{T}}.
\end{equation}
This formula isolates the two ways a sample moves when the sphere center $\mathbf{p}$ translates:
\begin{itemize}
    \item $\mathbf{I}_3$: the ray origin itself translates.
    \item $\mathbf{d}_b(\nabla_{\mathbf{p}}t_{s,k})^{\mathsf{T}}$: the sample distance changes as the voxel entry and exit distances change.
\end{itemize}

\subsubsection{Derivative of the Trilinear Raw Density}
Let $l_v$ be the side length of voxel $v$.
The voxel-local coordinates corresponding to a world-space point and sample $k$ are
\begin{equation}
    \boldsymbol{\xi}=\frac{\mathbf{x}-\mathbf{m}_v}{l_v},
    \qquad
    \boldsymbol{\xi}_{s,k}=\frac{\mathbf{x}_{s,k}-\mathbf{m}_v}{l_v}.
    \label{eq:normalized_voxel_coordinate}
\end{equation}

Let $Z_v$ denote the eight raw corner-density values of voxel $v$.
The sampled raw density is $Z_{s,k}=Z(\mathbf{x}_{s,k})=\operatorname{interp}(Z_v,\boldsymbol{\xi}_{s,k})$, where $\operatorname{interp}$ denotes trilinear interpolation of the corner values.
Its world-space spatial gradient is
\begin{equation}
    \nabla_{\mathbf{x}}Z_{s,k}
    =\frac{1}{l_v}
    \left.\nabla_{\boldsymbol{\xi}}\operatorname{interp}(Z_v,\boldsymbol{\xi})
    \right|_{\boldsymbol{\xi}=\boldsymbol{\xi}_{s,k}}.
\end{equation}
The derivative of the sampled raw density with respect to the sphere center is therefore
\begin{equation}
    \nabla_{\mathbf{p}}Z_{s,k}
    =\left(\frac{\partial\mathbf{x}_{s,k}}{\partial\mathbf{p}}\right)^{\mathsf{T}}
    \nabla_{\mathbf{x}}Z_{s,k}.
\end{equation}
Hence,
\begin{equation}
\begin{aligned}
    \nabla_{\mathbf{p}}Z_{s,k}
    &=\nabla_{\mathbf{x}}Z_{s,k} \\
    &\quad {}
    +(\nabla_{\mathbf{x}}Z_{s,k}\cdot\mathbf{d}_b)\nabla_{\mathbf{p}}t_{s,k}.
\end{aligned}
\end{equation}
The first term is the density-field gradient under direct translation of the ray origin.
The second term accounts for the midpoint sample moving along the ray as the intersection interval changes.

Let $\sigma_{s,k}=\explin(Z_{s,k})$.
Using the activation in \eqref{eq:explin},
\begin{equation}
\begin{aligned}
    \nabla_{\mathbf{p}}\sigma_{s,k}
    &=\explin'(Z_{s,k})\Bigl[
    \nabla_{\mathbf{x}}Z_{s,k} \\
    &\qquad {}
    +(\nabla_{\mathbf{x}}Z_{s,k}\cdot\mathbf{d}_b)\nabla_{\mathbf{p}}t_{s,k}
    \Bigr].
\end{aligned}
\end{equation}

\subsubsection{Derivative of Sampled Integrated Density}
For segment $s$, the sampled integrated density in \eqref{eq:sampled_segment_integral} is
\begin{equation}
    \widehat{\tau}_s=h_s\sum_{k=1}^{K_s}\sigma_{s,k}.
\end{equation}
The product rule gives
\begin{equation}
\begin{aligned}
    \nabla_{\mathbf{p}}\widehat{\tau}_s
    &=\left(\nabla_{\mathbf{p}}h_s\right)
    \sum_{k=1}^{K_s}\sigma_{s,k} \\
    &\quad {}
    +h_s\sum_{k=1}^{K_s}\nabla_{\mathbf{p}}\sigma_{s,k}.
\end{aligned}
\end{equation}
\begin{proof}[Proof of Theorem~\ref{thm:voxel_center_derivative}]
Substituting the spacing gradient $\nabla_{\mathbf{p}}h_s=(\gamma_s^{\mathrm{out}}-\gamma_s^{\mathrm{in}})/K_s$, the spacing $h_s=\ell_s/K_s$, and the density gradients derived above into this product rule yields \eqref{eq:voxel_center_derivative}.
\end{proof}

Theorem~\ref{thm:voxel_center_derivative} has a useful interpretation:
\begin{itemize}
    \item \textbf{interval term:}
    $(\gamma_s^{\mathrm{out}}-\gamma_s^{\mathrm{in}})[\cdots]$
    differentiates the length of the ray segment inside the voxel;
    \item \textbf{field-sampling term:}
    $\ell_s[\cdots]$
    differentiates the density values at the moving midpoint samples.
\end{itemize}
This separation is the discrete midpoint-quadrature analogue of differentiating an integral with moving limits.

\subsection{Pixel-Opacity Gradient}
\label{app:pixel_opacity_gradient}
Differentiating the rendered pixel-opacity relation in \eqref{eq:pixel_segment_opacity} gives the following result.
\begin{corollary}[Center to spherical-pixel opacity]
\label{cor:pixel_opacity_gradient}
For a spherical camera centered at $\mathbf{p}$ and a fixed pixel direction $\mathbf{d}_b$, the rendered pixel opacity satisfies
\[
    \nabla_{\mathbf{p}}A_b
    =(1-A_b)
    \sum_{s\in\mathcal{S}(\mathbf{d}_b)}
    \nabla_{\mathbf{p}}\widehat{\tau}_s,
\]
where $\nabla_{\mathbf{p}}\widehat{\tau}_s$ is given by Theorem~\ref{thm:voxel_center_derivative}.
The active ray--voxel intersection structure is held fixed during differentiation.
\end{corollary}

%% file: tex/pseudocode_ray_lipschitz.tex
Algorithm~\ref{alg:gpu_ray_lipschitz} summarizes the GPU evaluation of the ray error bound in \eqref{eq:per_ray_upper_bound}.
Each GPU thread evaluates one pixel ray using the ray--voxel intersection intervals computed by SVRaster's \textsc{RayAABB}~\cite{svraster}.
For each segment $s$, \textsc{RayCubicCoefficients} substitutes the parameterization in \eqref{eq:segment_parameterization} into the voxel's trilinear raw density to obtain the coefficients in \eqref{eq:ray_cubic}.
The maximum derivative magnitude is then found from the endpoints $t_s^{\mathrm{in}}$, $t_s^{\mathrm{out}}$ and any interior vertex, as in \eqref{eq:tight_ray_lipschitz}--\eqref{eq:ray_lipschitz_candidates}.
By \eqref{eq:explin_gradient_bound}, the resulting $L_{\mathrm{ray},s}$ also bounds the activated density along the segment.

\begin{algorithm}[!t]
\caption{Per-Ray Lipschitz Error Bound}
\label{alg:gpu_ray_lipschitz}
\footnotesize
\begin{algorithmic}[1]
\Require Sphere center $\mathbf{p}$ and unit pixel directions $\{\mathbf{d}_b\}_{b=1}^{B}$
\Require Intersected voxels for each pixel
\Require Voxel geometry and raw corner densities for each segment; sample counts $K_s\geq1$
\Ensure Per-pixel error bound $\varepsilon_{\mathrm{ray},b}$
\ForAll{pixels $b$ \textbf{in parallel}}
    \State $\varepsilon_{\mathrm{ray},b}\gets0$
    \ForAll{segments $s\in\mathcal{S}(\mathbf{d}_b)$ with intersecting voxel $v$}
        \State $(t_s^{\mathrm{in}},t_s^{\mathrm{out}})\gets\Call{RayAABB}{\mathbf{p},\mathbf{d}_b,v}$
        \State $\ell_s\gets t_s^{\mathrm{out}}-t_s^{\mathrm{in}}$
        \State $(c_{s,1},c_{s,2},c_{s,3})\gets\Call{RayCubicCoefficients}{v,s}$
        \State $L_{\mathrm{ray},s}\gets\max\bigl(|p_s'(t_s^{\mathrm{in}})|,|p_s'(t_s^{\mathrm{out}})|\bigr)$
            \Comment{$p_s'$ in \eqref{eq:ray_cubic_derivative}}
        \If{$c_{s,3}\neq0$}
            \State $t^{\star}\gets-c_{s,2}/(3c_{s,3})$
            \If{$t_s^{\mathrm{in}}\leq t^{\star}\leq t_s^{\mathrm{out}}$}
                \State $L_{\mathrm{ray},s}\gets\max\bigl(L_{\mathrm{ray},s},|p_s'(t^{\star})|\bigr)$
            \EndIf
        \EndIf
        \State $\varepsilon_{\mathrm{ray},b}\gets\varepsilon_{\mathrm{ray},b}
            +L_{\mathrm{ray},s}\ell_s^2/(2K_s)$
    \EndFor
\EndFor
\end{algorithmic}
\end{algorithm}

The resulting ray error bound $\varepsilon_{\mathrm{ray},b}$ and the angular error bound $\varepsilon_{\mathrm{ang},b}$ from Algorithm~\ref{alg:gpu_angular_error} form the total error bound $\varepsilon_b=\varepsilon_{\mathrm{ray},b}+\varepsilon_{\mathrm{ang},b}$, which is combined with the pixel opacities $A_b$ in \eqref{eq:rendered_collision_upper}.

%% file: tex/pseudocode_angular_error.tex
Algorithm~\ref{alg:gpu_angular_error} summarizes the GPU evaluation of the angular error bound.
Angular bins $\mathcal{B}_b$ are processed in parallel, with one GPU block per bin.
Within each block, threads process in parallel the voxels $v\in\mathcal{V}_b$ that lie in the region covered by angular bin $\mathcal{B}_b$, using the precomputed bounds $L_{\mathrm{vox},v}$ in \eqref{eq:voxel_lipschitz}.
Each voxel thread traverses the radial shells sequentially and atomically updates $G_{b,j}$ for every intersected region.
We use uniform shell boundaries $r_j=jR/N_r$, a special case of the radial partition in Section~\ref{subsec:angular_bound}.

\begin{algorithm}[!t]
\caption{Angular Error Bound from Radial-Shell Lipschitz Maxima}
\label{alg:gpu_angular_error}
\footnotesize
\begin{algorithmic}[1]
\Require Radius $R>0$, angular radius $\delta_b=\delta\geq0$, and shell count $N_r\geq1$
\Require Intersecting voxel sets $\mathcal{V}_b$ for $b=1,\ldots,B$
\Require Chord-swept regions $\widetilde{\mathcal{F}}_{b,j}$ in \eqref{eq:chord_swept_region} and bounds $L_{\mathrm{vox},v}\geq0$
\Ensure Angular error bounds $\varepsilon_{\mathrm{ang},b}$ for $b=1,\ldots,B$
\State $r_j\gets jR/N_r$ for $j=0,\ldots,N_r$
\ForAll{angular bins $b$ \textbf{in parallel}}
    \State $G_{b,j}\gets0$ for $j=0,\ldots,N_r-1$
    \ForAll{voxels $v\in\mathcal{V}_b$ \textbf{in parallel}}
        \ForAll{radial shells $j=0,\ldots,N_r-1$}
            \If{voxel $v$ intersects $\widetilde{\mathcal{F}}_{b,j}$}
                \State \Call{AtomicMax}{$G_{b,j},L_{\mathrm{vox},v}$}
            \EndIf
        \EndFor
    \EndFor
    \State $\varepsilon_{\mathrm{ang},b}\gets2\sin\!\left(\frac{\delta_b}{2}\right)\sum_{j=0}^{N_r-1}G_{b,j}\frac{r_{j+1}^2-r_j^2}{2}$
\EndFor
\end{algorithmic}
\end{algorithm}

The final sum is evaluated only after all voxel threads in the block have finished updating $G_{b,j}$.